\documentclass[11pt,a4paper]{article}

\usepackage[numbers,sort&compress]{natbib}
\usepackage[ruled,vlined,linesnumbered]{algorithm2e}
\usepackage{xcolor}
\usepackage{enumitem}
\usepackage{amssymb,amsthm,amsmath,bm}
\usepackage{graphicx,booktabs,multirow,diagbox}
\usepackage{subfigure}
\usepackage{caption}

\usepackage{hyperref}
\usepackage[a4paper,margin=1in]{geometry}
\usepackage{placeins}

\title{Towards Automated Knowledge Transfer in Evolutionary Multitasking via Large Language Models}

\author{
\normalsize
Xuebin Lyu$^{1}$, Yuxiao Huang$^{2}$, Xuefeng Chen$^{1}$, Jing Tang$^{3}$,\\
Liang Feng$^{1}$\thanks{Corresponding author: liangf@cqu.edu.cn}, Kay Chen Tan$^{4}$\\[1ex]
\small
\parbox{0.9\textwidth}{\centering
$^{1}$College of Computer Science, Chongqing University, Chongqing 400044, China\\
$^{2}$Department of Information Technology, Chongqing Ant Consumer Finance Co., Ltd, Chongqing 400044, China\\
$^{3}$Data Science and Analytics Thrust, The Hong Kong University of Science and Technology (Guangzhou), Guangzhou 511453, China\\
$^{4}$Department of Data Science and Artificial Intelligence, The Hong Kong Polytechnic University, Hong Kong SAR 999077, China
}
}

\date{}

\begin{document}

\maketitle

\begin{abstract}
Evolutionary multi-task optimization (EMTO) is an advanced optimization paradigm that improves search efficiency by enabling knowledge transfer across multiple tasks solved in parallel. Accordingly, a broad range of knowledge transfer methods (KTMs) have been developed as integral components of EMTO algorithms, most of which are tailored to specific problem settings. However, the design of effective KTMs typically relies on substantial domain expertise and careful manual customization, as different EMTO scenarios require distinct transfer strategies to achieve performance gains. Meanwhile, recent advances in large language models (LLMs) have demonstrated strong capabilities in autonomous programming and algorithm synthesis, opening up new possibilities for automating the design of optimization solvers. Motivated by this, in this paper, we propose a Self-guided Knowledge Transfer Design (SKTD) framework that leverages LLMs to autonomously generate knowledge transfer methods (KTMs) as algorithmic components within EMTO. By enabling data-driven and self-adaptive construction of transfer strategies, SKTD facilitates effective knowledge reuse across heterogeneous tasks and diverse EMTO scenarios. To the best of our knowledge, this work represents the first attempt to automate the generation of KTMs for EMTO. Extensive experiments on well-established EMTO benchmarks with varying degrees of task similarity demonstrate that the proposed SKTD consistently achieves superior or highly competitive performance compared with both the state-of-the-art program search approach and manually designed EMTO methods, in terms of optimization effectiveness and cross-scenario generalization.

\end{abstract}


\section{Introduction}

multi-task optimization (EMTO) integrates evolutionary optimization with knowledge transfer \cite{gupta2015multifactorial,tan2021evolutionary,osaba2022evolutionary}, providing a unified framework for simultaneously solving multiple optimization tasks by exploiting shared information generated during the evolutionary search process.
In this framework, shared knowledge is typically extracted through knowledge transfer methods (KTMs) derived from problem characteristics or accumulated search experiences, and plays a critical role in improving search efficiency and accelerating convergence toward high-quality solutions across related tasks \cite{zhao2023makes,xue2023solution}. Over the past decade, EMTO has demonstrated notable success in a broad range of real-world optimization scenarios, including complex engineering design \cite{gupta2022half}, feature selection \cite{yang2025evolutionary}, robotic arm control \cite{wu2023transferable}, and online price promotion \cite{huang2023evolutionary}. 
These achievements reflect sustained methodological progress in KTMs, which have evolved from early implicit transfer strategies to explicit transfer mechanisms, and have been further strengthened through the incorporation of neural network-based models \cite{tan2023knowledge}.
Despite these advances, the effectiveness of KTMs remains highly scenario-dependent, as discrepancies between task characteristics and the underlying assumptions of transfer mechanisms could limit or even degrade optimization performance.
Moreover, the design of robust and effective KTMs continues to rely heavily on domain-specific expertise and extensive manual effort, resulting in substantial human resource costs and limited scalability across diverse EMTO settings.

In recent years, advances in the problem-solving capabilities of large language models (LLMs) have stimulated growing research interest in their application to optimization problems \cite{wu2024evolutionary,zhang2025systematic}.
Early studies primarily explored the direct use of LLMs as search operators, including combinatorial optimizers for the traveling salesman problem \cite{huang2025evaluation,liu2024large}, crossover operators for black-box optimization \cite{lange2024large}, and evolutionary optimizers for numerical optimization \cite{liu2025large}. 
However, such approaches tend to underperform on numerical optimization tasks once problem scales exceed small instances. This limitation largely arises because LLMs are trained for token-level text generation rather than numerical computation, which restricts their ability to effectively handle floating-point arithmetic and high-dimensional search spaces \cite{huang2024exploring}.

More recent research has shifted toward leveraging LLMs as generators of optimization algorithms instead of as direct search operators.
By exploiting their strong text processing and code generation capabilities, these approaches employ LLMs for program evolution to automatically synthesize problem-specific solvers, enabling their application to more complex optimization tasks.
Empirical studies have demonstrated the effectiveness of this paradigm in cost-aware Bayesian optimization \cite{yao2024evolve}, photonic structure optimization \cite{yin2025optimizing}, and multi-objective optimization \cite{huang2025autonomous}.
In light of these developments, the algorithm generation capability of LLMs presents a natural opportunity to automate the construction of knowledge transfer methods that adapt to problem characteristics in EMTO. 
To the best of our knowledge, autonomous KTM generation has not yet been explored in evolutionary multitasking, and this work therefore represents an initial step toward addressing this open problem.

Building on these observations, this paper proposes an LLM-assisted knowledge transfer design method for evolutionary multitasking that autonomously explores and generates KTMs for diverse EMTO scenarios. The proposed approach leverages LLM-based search operators to initialize and iteratively evolve KTMs within a cyclic search framework. 
To further enhance the effectiveness of KTM evolution, a self-guided knowledge extraction mechanism is developed to distill both fine-grained and coarse-grained design insights accumulated throughout the search process, which are then used to guide the LLM in refining newly generated KTMs. 
Unlike existing LLM-based autonomous programming approaches that primarily rely on recombining or synthesizing existing programs, the proposed method explicitly exploits experiential feedback collected during the search process, enabling the generation of more informative and adaptive KTM designs.
Extensive empirical evaluations demonstrate that the proposed method achieves superior or competitive optimization performance compared with both the state-of-the-art program search method and manually designed EMTO counterparts.
The main contributions of this work are summarized as follows:

\begin{itemize}
  \item First, the design of knowledge transfer methods in evolutionary multitasking is formulated as an automated search problem, in which LLM is adopted to generate and refine knowledge transfer strategies, replacing manual and scenario-specific design practices.
  \item Second, an integrated EMTO paradigm is developed that incorporates LLM-driven program synthesis as a core component, enabling the systematic construction of scenario-adaptive knowledge transfer methods.
  \item Third, a self-guided knowledge utilization mechanism is introduced to extract structured design information from search trajectories and to incorporate this information into subsequent KTM generation, enabling experience-informed refinement of knowledge transfer strategies.
\end{itemize}

The rest of this paper is organized as follows. Section \ref{sec:literature} first reviews existing work on KTMs in EMTO, and then introduces LLM-driven autonomous programming for optimization. Our proposed method, which is able to generate and explore novel KTMs in a given EMTO scenario, is detailed in Section \ref{sec:methodology}. Next, comprehensive empirical studies are conducted in Section \ref{sec:experiment} to validate the performance of the proposed method. Lastly, Section \ref{sec:conclusion} concludes this work and discusses several potential directions for future research.

\section{Preliminary} \label{sec:literature}
This section reviews representative KTMs in EMTO to clarify their underlying principles and limitations. It then introduces LLM-based autonomous programming techniques that form the methodological foundation of the proposed approach.

\subsection{Knowledge Transfer in EMTO}

\begin{figure}[htbp]
  \centering
  \includegraphics[width=0.5\columnwidth]{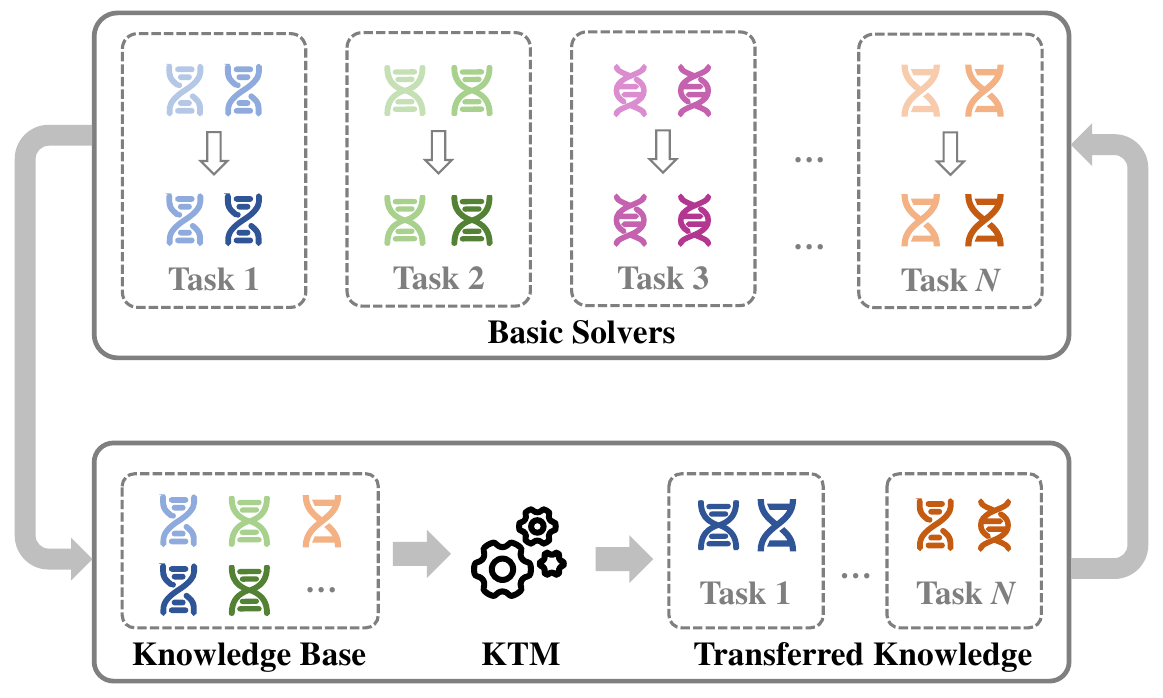}
  \caption{Illustration of the evolutionary multi-task optimization.}
  \label{fig:emto}
\end{figure}

A representative illustration of the EMTO paradigm is shown in Fig. \ref{fig:emto}, in which multiple basic solvers address distinct optimization tasks in parallel, while the KTM enables inter-task knowledge sharing to guide and enhance their search processes.
Within EMTO, the KTM maintains a collective knowledge base constructed from information gathered across all tasks and produces task-specific transferred knowledge to support individual solvers.
In this context, the KTM determines both the form of knowledge representation and the transfer mechanism, thereby addressing the fundamental questions of what knowledge to transfer and how to transfer it as evolutionary multitasking proceeds online.
Consequently, the selection and design of an appropriate KTM play a crucial role in improving overall optimization performance in EMTO \cite{wei2021review,tan2023knowledge}.

\begin{figure}[htbp]
\centering
\includegraphics[width=0.5\columnwidth]{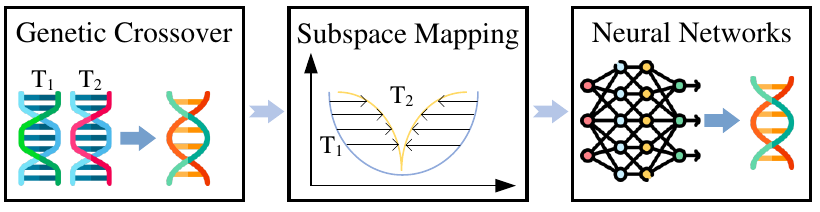}
\caption{Development of knowledge transfer methods in EMTO.}
\label{fig:emtoevolv}
\end{figure}

To promote positive knowledge transfer, the design of KTMs has evolved continuously to accommodate the diverse characteristics of EMTO scenarios, as illustrated in Fig. \ref{fig:emtoevolv}. Early studies predominantly adopted genetic crossover as the KTM \cite{gupta2015multifactorial,VC,ding2017generalized,gupta2016multiobjective,bali2019multifactorial}. These approaches typically require a unified solution representation across all tasks, with knowledge transfer realized by performing crossover between solutions from different tasks. While computationally efficient, such mechanisms impose strong task similarity assumptions and often fail to achieve satisfactory performance when these assumptions are violated. To mitigate this limitation, subspace mapping-based KTMs were subsequently proposed. In these methods, each task is projected onto an individual subspace, and knowledge transfer is enabled through learned mappings between source and target subspaces at the representation level \cite{AE,AF,feng2020explicit}. 
This line of work generally depends on prior task analysis to establish pairwise task relationships, and the lightweight learning models employed may be insufficient to capture complex inter-task structures in challenging optimization problems. 

More recently, neural network-based KTMs have been introduced as powerful knowledge learning and transfer modules, enabling effective many-task optimization \cite{huang2023evolutionary,dai2024multi,feng2025lifelong}. Despite their flexibility, such KTMs incur substantial computational overhead and may be unnecessary in scenarios where simpler transfer mechanisms are sufficient to achieve competitive performance.
In addition, the training of neural network-based models typically requires large amounts of data \cite{cheng1994neural}, which limits their applicability in data-scarce EMTO settings.
Taken together, these developments indicate that no single KTM is universally effective across all EMTO scenarios, and that optimization performance critically depends on the alignment between task characteristics and the underlying transfer mechanism. However, the design of scenario-appropriate KTMs continues to rely heavily on domain-specific expertise, motivating the need for a systematic and automated approach to KTM construction tailored to diverse EMTO scenarios.
\subsection{Autonomous Programming via LLM}

Large language models (LLMs) are a class of deep neural networks trained on large-scale text corpora that exhibit strong capabilities in natural language understanding and generation.
Recent advances have demonstrated their effectiveness in a wide range of application domains, including engineering design \cite{wong2025llm2tea}, recommender systems \cite{liu2025language}, spatio-temporal forecasting \cite{li2025causal}, and automatic academic paper rating \cite{liu2025lmcbert}.
With continued progress in model architectures and training strategies, LLMs have also shown increasing proficiency in program synthesis, which has significantly advanced research in autonomous programming \cite{weng2020autonomous}. 
In the context of optimization, autonomous programming focuses on automatically generating executable optimizer code to solve complex optimization problems, and existing approaches can be broadly categorized into universal and domain-specific methods \cite{wu2024evolutionary}.
Universal methods aim to develop solvers with strong cross-domain generality that perform robustly across diverse problem settings \cite{liventsev2023fully,xu2025evospeak,zhang2025llm}.
However, such methods often struggle to effectively address complex or highly specialized tasks, and tend to underperform on black-box problems that are common in real-world optimization scenarios \cite{liventsev2023fully,huang2025autonomous}.

In contrast, domain-specific methods aim to efficiently solve particular classes of optimization problems by tailoring optimizer generation to the intrinsic properties of individual tasks, thereby offering a promising direction for performance improvement in specialized scenarios \cite{wu2024evolutionary}.
Recent studies have demonstrated the potential of LLMs in discovering mathematical solutions through program search, highlighting their capability to address challenging problems such as the cap set problem \cite{romera2024mathematical}. 
In addition, LLMs have been integrated as hyper-heuristics to facilitate the automated problem formulation and heuristic design for combinatorial optimization tasks \cite{zhang2025agentic,ye2024reevo}.
Further progress has been reported in autonomous solver design for multi-objective optimization, where LLM-driven approaches have achieved superior performance compared with conventional methods \cite{huang2025autonomous}.

Despite these advances, domain-specific autonomous programming has primarily focused on the design of base-level optimizers, while higher-level algorithmic components remain largely unexplored in evolutionary multitasking.
Such components, including knowledge transfer mechanisms, are typically less structured and more flexible than task-level optimizers. Their effective design therefore requires strong exploratory capability to identify promising mechanisms, together with effective exploitation of accumulated experience to progressively refine them.

\section{Proposed Method} \label{sec:methodology}

This section presents the proposed Self-guided Knowledge Transfer Design (SKTD) algorithm, which employs LLMs to autonomously explore effective knowledge transfer methods for diverse evolutionary multi-task optimization scenarios. It first provides an overview of the SKTD framework, followed by a detailed description of its three core components.

\subsection{Overview of The SKTD Framework} \label{subsec:overview}

\begin{figure}[!b]
	\centering
	\includegraphics[width=0.999\columnwidth]{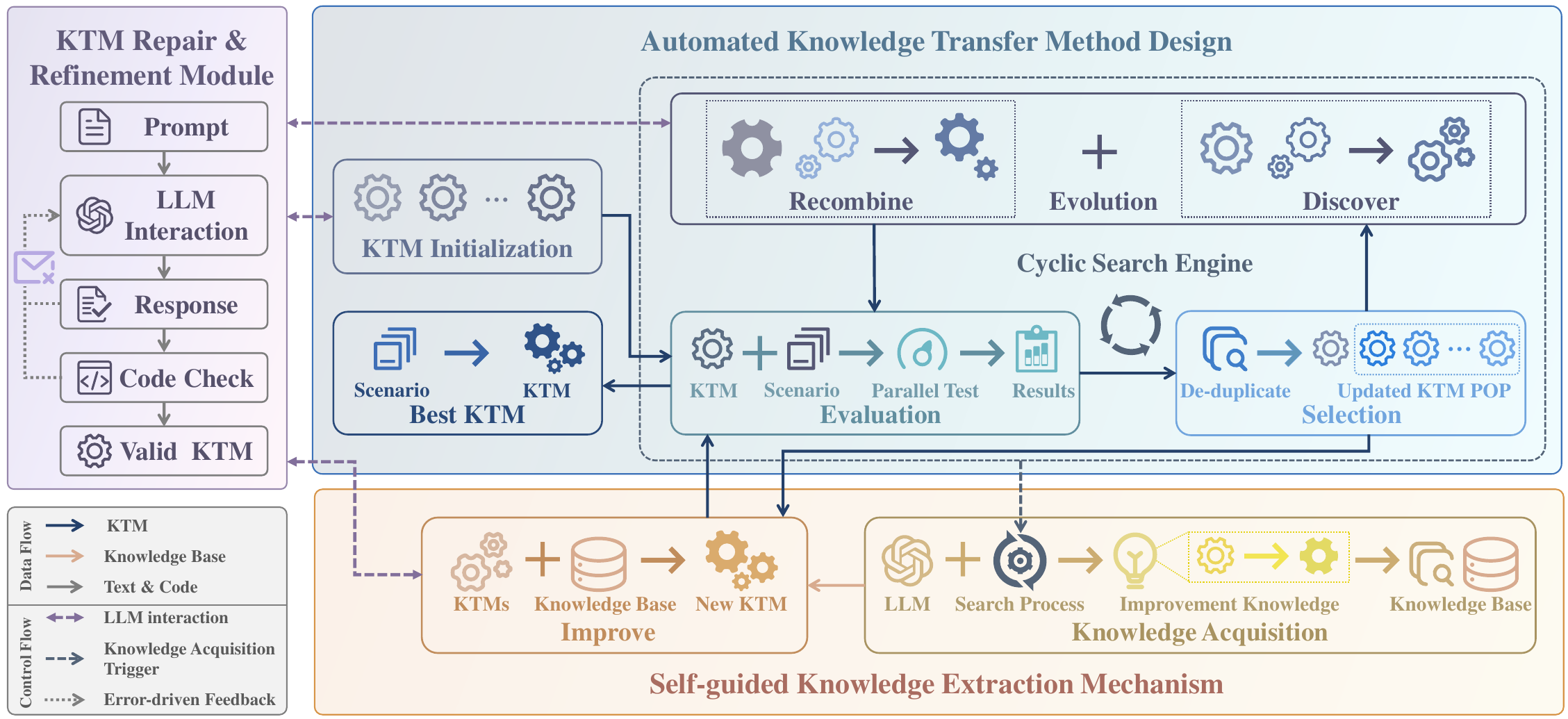}
	\caption{Illustration of the proposed SKTD algorithm.}
	\label{fig:workflow}
\end{figure}

An overview of SKTD is shown in Fig. \ref{fig:workflow}, which illustrates its three tightly integrated components: automated KTM design, a self-guided knowledge extraction mechanism, and a KTM repair and refinement module. 
Automated KTM design forms the core of the framework. It initializes a population of candidate KTMs and iteratively explores transfer strategies through a cyclic search process tailored to the current EMTO scenario.
During this process, the self-guided knowledge extraction mechanism utilizes the LLM to analyze search trajectories and to distill structured design feedback, such as recurring deficiencies, effective transfer patterns, and task-specific adaptation behaviors, which is then used to inform subsequent KTM construction.
To ensure practical usability, the KTM repair and refinement module transforms LLM-generated KTM programs into executable implementations by enforcing interface consistency, correcting syntactic and runtime errors, and completing missing functional components, thereby enabling reliable evaluation and iterative improvement of candidate KTMs.

The overall procedure of the proposed SKTD algorithm is summarized in Alg. \ref{alg:all}. The algorithm begins with the automated KTM design component by initializing a population of candidate knowledge transfer methods, denoted as $\mathcal{P}_{ktm}$. 
These KTMs are first evaluated under the given EMTO scenario, as indicated in line 1, with the evaluation procedure described in Section \ref{subsec:exp_set}. 
The population $\mathcal{P}_{ktm}$ is then iteratively evolved through a cyclic search process, in which new KTMs, denoted as \(\text{KTM}^{new}\), are generated from existing ones using multiple LLM-driven operators.
In particular, the \textit{recombine} operator (line 9) and the \textit{discover} operator (line 11) are probabilistically selected to explore new transfer strategies, corresponding to the exploration capability of the automated KTM design component.
In parallel, the self-guided knowledge extraction mechanism analyzes the historical evolution of KTMs during the search process and produces structured design feedback, denoted as \(\text{KLG}^{new}\).
When such feedback is available, it probabilistically activates the \textit{improve} operator (lines 5-6) to refine existing KTMs in a targeted manner, thereby enabling experience-informed exploitation.
Each extracted piece of knowledge is stored in an experiential knowledge base $\mathcal{K}$ to support subsequent KTM refinement (line 15).
To ensure executability and reliable evaluation, each newly generated KTM is processed by the KTM repair and refinement module before being evaluated within the EMTO scenario. 
The population $\mathcal{P}_{ktm}$ is then updated accordingly (line 13), following the procedure detailed in Section~\ref{subsec:automated}. 
This iterative process continues until a predefined stopping condition is met, yielding a scenario-adaptive KTM as the final output.
The following subsections describe the three components of SKTD in detail.

\begin{algorithm}[htbp]
	\caption{Pseudocode of the proposed SKTD algorithm.}
	\SetArgSty{textnormal}
	\label{alg:all}
	\KwIn{
		\\\quad \(G_{ev}\): Number of generations to evolve KTMs.
		\\\quad \(N_{ktm}\): Size of the KTM population.
		\\\quad \(PBs\): The problems to be solved.
	}
	\KwOut{
		\\\quad \(\text{KTM}^{*}\): The best KTM.
	}
	Initialize $\mathcal{P}_{ktm}$ and evaluate it on $PBs$.\\
	\(\mathcal{K} \leftarrow \{\}\) \\
	\For{$gen \leftarrow 1$ to $G_{ev}$}{
	\For{$i \leftarrow 1$ to $N_{ktm}$}{
		\If{$\mathcal{K} \neq \emptyset$ \textbf{and} $rand() < p_{imp}$}{
			Obtain $\text{KTM}^{new}$ via \textit{improve} operator in Alg. \ref{alg:improve}.\\
		}
		\uElseIf{$rand() < p_{rcb}$}{
			Obtain $\text{KTM}^{new}$ via \textit{recombine} operator.\\
	
		}
		\Else{
			Obtain $\text{KTM}^{new}$ via \textit{discover} operator.\\
	
		}
		Evaluate $\text{KTM}^{new}$ and update $\mathcal{P}_{ktm}$.\\
		\If{Knowledge extraction is triggered}{
			Obtain $\text{KLG}^{new}$ and update $\mathcal{K}$.\\
		}
	}}
	
\end{algorithm}

\subsection{Automated KTM Design} \label{subsec:automated}

This subsection describes the automated KTM design pipeline underlying the proposed method.
As illustrated in Fig. \ref{fig:workflow}, the pipeline begins with the construction of an initial population of candidate knowledge transfer methods, followed by a cyclic search process that systematically explores and refines KTMs for a given multi-task optimization scenario. 
In each iteration, newly generated KTMs are evaluated within the current EMTO setting to provide feedback for population update and subsequent search.
These evaluated KTMs are incorporated into the population through a selection mechanism that regulates both solution quality and population diversity.
To this end, a de-duplication step is first introduced to identify and discard redundant KTMs whose behaviors or performance characteristics suggest substantial similarity to existing population members.
Candidate KTMs that pass this filtering step are then compared against the current population and admitted by replacing inferior designs when improvement is observed.
The resulting updated population is subsequently used to drive the next round of KTM generation, forming a closed-loop optimization process.
The key stages of this pipeline, including KTM initialization and KTM evolution, are detailed in the following subsections.

\subsubsection{KTM Initialization}

In the initialization stage of the proposed method, the LLM is leveraged to generate KTMs by responding to an informative prompt, as depicted in Fig. \ref{prompt:initial}.
As shown, the prompts consist of two parts: the system section, which serves as the system prompt, and the description section, which functions as the user prompt \cite{schulhoff2024prompt}.
The system section configures the LLM's role and outlines the task context, thereby guiding the model toward generating accurate and relevant responses.
The description section defines both the objective and the detailed specifications for KTM generation.
Given that LLMs may possess limited understanding of the EMTO paradigm, they are instructed to draw insights from existing evolutionary transfer optimization methods reported in the literature, potentially contributing to the generation of high-quality KTMs \cite{wei2022chain}.
Within this prompt, the placeholders \#PROBLEM\# and \#PROBLEM\_DESC\# represent the name of the problem and its informative description.
Additionally, \#FORMAT\# defines the unified input-output interface of KTMs, enabling the generated KTMs to seamlessly integrate into and operate within the EMTO framework.
The initialization stage is conducted through iterative interactions with the LLM and continues until the number of generated KTMs reaches the population size.

\begin{figure}[htbp]
  \centering
  \fcolorbox{black}{gray!10}{\parbox{0.9\linewidth}{
  \begin{center}
  \textcolor[rgb]{0.0,0.0,0.0}{\textbf{Initialization}}\\
  \end{center}
  \textcolor[rgb]{0.604, 0.137, 0.541}{\textbf{\textit{System}:}
  You are an expert in designing powerful evolutionary transfer optimization methods which enable effective and efficient knowledge transfer to facilitate the optimization of multiple \#PROBLEM\#. \#PROBLEM\_DESC\#.\\
  \textbf{\textit{Description}:}
  Given multiple \#PROBLEM\#, you are required to design an innovative evolutionary transfer optimization algorithm using Python code for enhanced optimization performance across \#PROBLEM\#.
  \textbf{Drawing inspiration from existing evolutionary transfer optimization methods in the literature}, the qualified solutions within one \#PROBLEM\# can be transformed and transferred to another, thereby enhancing the search performance for each \#PROBLEM\#.\\
  Help me implement an evolutionary transfer optimization method or model in Python as a function named LLMTransfer. Below is the specified format for the function you need to design. \\
  \#FORMAT\# \\
  Do not give additional explanations. \\
  }
  }}
  \captionof{figure}{Prompt template for the \textit{initialize} operator}
  \label{prompt:initial}
\end{figure}

\subsubsection{KTM Evolution}

\begin{figure}[htbp]
  \centering
  \includegraphics[width=0.55\columnwidth]{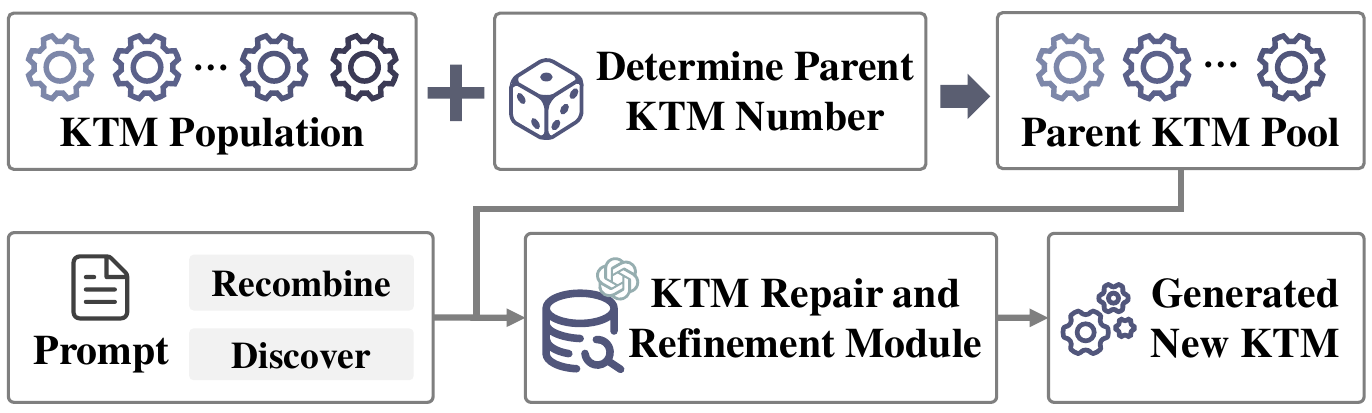}
  \caption{Illustration of the KTM evolution stage.}
  \label{fig:evo}
\end{figure}

In the evolution stage of the proposed SKTD algorithm, each new KTM is generated from the selected parent KTMs, as illustrated in Fig.~\ref{fig:evo}.
The number of parent KTMs, denoted by $N_{evo}$, is dynamically adjusted to promote the diversity of the offspring KTMs during the search process \cite{huang2025autonomous}. Specifically, $N_{evo}$ is sampled as a random integer satisfying $2\leq N_{evo} < \left\lceil  N /2  \right\rceil$, where $N$ denotes the size of the KTM population.
Subsequently, $N_{evo}$ KTMs are selected using the roulette wheel selection method \cite{katoch2021review} to construct the parent KTM pool.
These parent KTMs are then employed to generate a new offspring with the guidance of an operator-specific prompt.

By adopting dedicated operator-specific prompts, SKTD supports both the \textit{recombine} and the \textit{discover} operators.
The \textit{recombine} operator aims to explore promising combinations of previously identified knowledge transfer components, thereby synthesizing superior KTMs from existing building blocks. In contrast, the \textit{discover} operator promotes the creation of novel KTMs by seeking components or configurations that differ significantly from known knowledge transfer structures, thereby broadening the search space. 
Both the \textit{recombine} and the \textit{discover} operators are described in detail in what following.

\noindent$\bullet$ \textit{Recombine} Operator

The \textit{recombine} operator instructs the LLM to synthesize a new KTM by recombining a subset of previously evaluated candidates, playing a pivotal role in constructing high-quality KTMs after the initialization stage.
The overall structure of the \textit{recombine} prompt is presented in Fig. \ref{prompt:recombine}.
In this prompt, \#$N_{evo}$\# specifies the number of parent KTMs, whereas \#EVOList\# contains the corresponding code snippets.
In addition, the code format of KTMs is included in the prompt to mitigate potential syntactic and semantic errors.
By selecting different KTMs to form the parent KTM pool and applying the \textit{recombine} prompt, the \textit{recombine} operator facilitates the generation of diverse offspring KTMs.

\begin{figure}[htbp]
	\centering
	\fcolorbox{black}{gray!10}{\parbox{0.9\linewidth}{
	\begin{center}
	\textcolor[rgb]{0.0,0.0,0.0}{\textbf{Recombine}}\\
	\end{center}
	\textcolor[rgb]{0.0,0.55,0.55}{\textbf{\textit{Description}:}
	I will demonstrate several effective and efficient LLMTransfer functions in XML format. Your task is to conceive a novel function with the same input/output formats and functionality, termed `LLMTransfer', which should \textbf{recombine useful sub-modules within these LLMTransfer functions} given to you. Below, you will find the \#$N_{evo}$\# LLMTransfer functions. \\
	\#EVOList\# \\
	Below is the specified format for the function you need to design. \\
	\#FORMAT\# \\
	No Explanation Needed!! \\
	}
	}}
	\captionof{figure}{Prompt template for the \textit{recombine} operator}
	\label{prompt:recombine}
\end{figure}

\noindent$\bullet$ \textit{Discover} Operator

The \textit{discover} operator is designed to foster innovation by promoting the exploration of novel KTM submodules that have not been previously encountered, thereby enhancing the diversity of the generated KTMs and expanding the search space beyond known configurations.
To mitigate potential performance degradation caused by exploratory modifications \cite{bao2022mutations}, the \textit{discover} prompt is crafted to guide the LLM to generate novel and effective functions, as illustrated in Fig. \ref{prompt:discover}.
Despite its structural similarity to the \textit{recombine} prompt, the \textit{discover} prompt differs fundamentally in its intent: the \textit{recombine} prompt instructs the LLM to ``recombine useful submodules within these LLMTransfer functions'', while the \textit{discover} prompt focuses on ``designing an innovative function distinct from all the provided ones''.

\begin{figure}[htbp]
  \centering
  \fcolorbox{black}{gray!10}{\parbox{0.9\linewidth}{
  \begin{center}
  \textcolor[rgb]{0.0,0.0,0.0}{\textbf{Discover}}\\
  \end{center}
  \textcolor[rgb]{0.0,0.4,0.7}{\textbf{\textit{Description}:}
  I will introduce several powerful evolutionary transfer function named LLMTransfer. Your task is to \textbf{design an innovative function that is distinct from all the provided ones}, while ensuring that the input/output formats, function name, and core functionality remain unchanged. Below, you will find the \#$N_{evo}$\# LLMTransfer functions. \\
  \#EVOList\# \\
  Below is the specified format for the function you need to design. \\
  \#FORMAT\# \\
  No Explanation Needed!! \\
  }
  }}
  \captionof{figure}{Prompt template for the \textit{discover} operator}
  \label{prompt:discover}
\end{figure}

\subsection{Self-guided Knowledge Extraction Mechanism}
Recognizing the substantial value of knowledge extracted from past experiences in strengthening optimization and learning capabilities \cite{gupta2017insights,shakya2023reinforcement}, the proposed method introduces a self-guided knowledge extraction mechanism. 
This mechanism produces structured design feedback by analyzing the historical evolution throughout the search process.
The evolution pipeline within the SKTD algorithm comprises two levels: the evolution of individual KTMs and that of the KTM population. 
These two levels of evolution correspond to individual- and population-level knowledge, respectively. 
To exploit the extracted knowledge, an \textit{improve} operator is incorporated to refine existing KTMs and guide the search in a strategic manner.
The individual- and population-level knowledge acquisition mechanisms and the \textit{improve} operator are detailed in the following.

\begin{figure}[b]
	\centering
	\includegraphics[width=0.55\columnwidth]{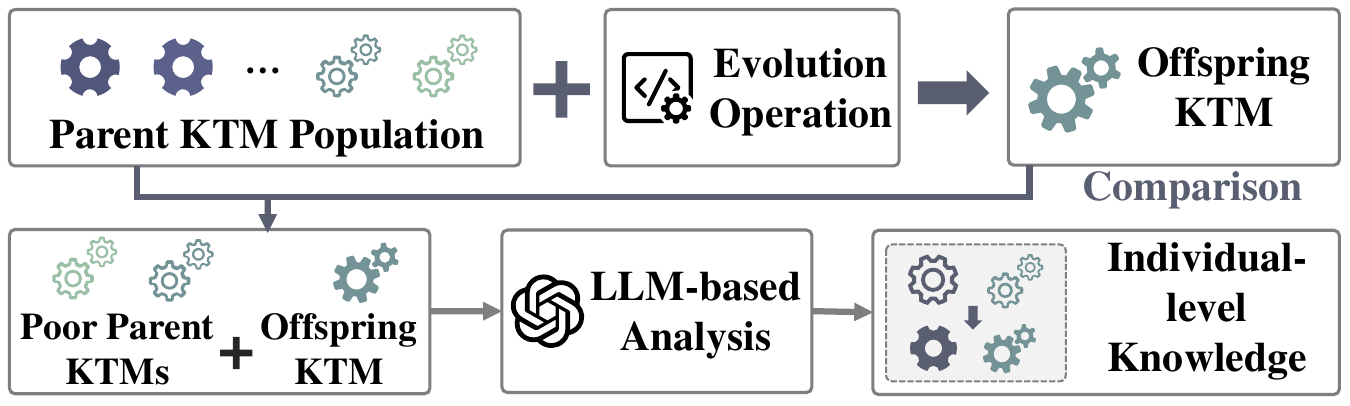}
	\caption{Illustration of the individual-level knowledge acquisition mechanism.}
	\label{fig:lka}
\end{figure}
\subsubsection{Individual-level Knowledge Acquisition}

The individual-level knowledge acquisition mechanism focuses on distilling actionable insights from parent-offspring transformations. 
As shown in Fig. \ref{fig:lka}, this mechanism examines the evolution process and identifies parent KTMs that underperform their offspring for contrastive analysis.
This analysis leverages prompt-driven interactions with the LLM to guide it to generate a concise summary of the findings, as illustrated in Fig. \ref{prompt:LKA}.
The placeholders \#HPKTM\# and \#UDKTM\# in the prompt denote the high-performing offspring KTM and the underperforming parent KTMs, respectively.
The placeholder \#KLG-XML-FORMAT\# specifies the format of the returned knowledge.
The individual-level knowledge acquisition mechanism is triggered whenever a newly generated offspring KTM outperforms any of its parent KTMs.

\begin{figure}[!t]
  \centering
  \fcolorbox{black}{gray!10}{\parbox{0.9\linewidth}{
  \begin{center}
  \textcolor[rgb]{0.0,0.0,0.0}{\textbf{Individual-level Knowledge Acquisition}}\\
  \end{center}
  \textcolor[rgb]{0.7, 0.3, 0.0}{\textbf{\textit{Description}:}
  You are provided with a high-performing evolutionary transfer optimization method implemented as a Python function named LLMTransfer, which has been developed by improving upon several underperforming approaches.\\
  Your task is to:\\
  1. Perform a step-by-step contrastive analysis between the high-performing LLMTransfer function and each of the poorer versions.\\
  2. Identify and explain key differences in algorithmic design, parameter tuning, transfer mechanisms, and performance strategies.\\
  3. Summarize actionable tips or ideas that could help improve the underperforming methods.\\
  High-performing method:
  \#HPKTM\#\\
  Underperforming methods:
  \#UDKTM\#\\
  Summarize your improvement tips or ideas in a single graph within 100 words based on the following structure:
  \#KLG-XML-FORMAT\#\\
  }
  }}
  \captionof{figure}{Prompt template for individual-level knowledge acquisition}
  \label{prompt:LKA}
  \end{figure}

\subsubsection{Population-level Knowledge Acquisition}

\begin{figure}[!b]
	\centering
	\includegraphics[width=0.55\columnwidth]{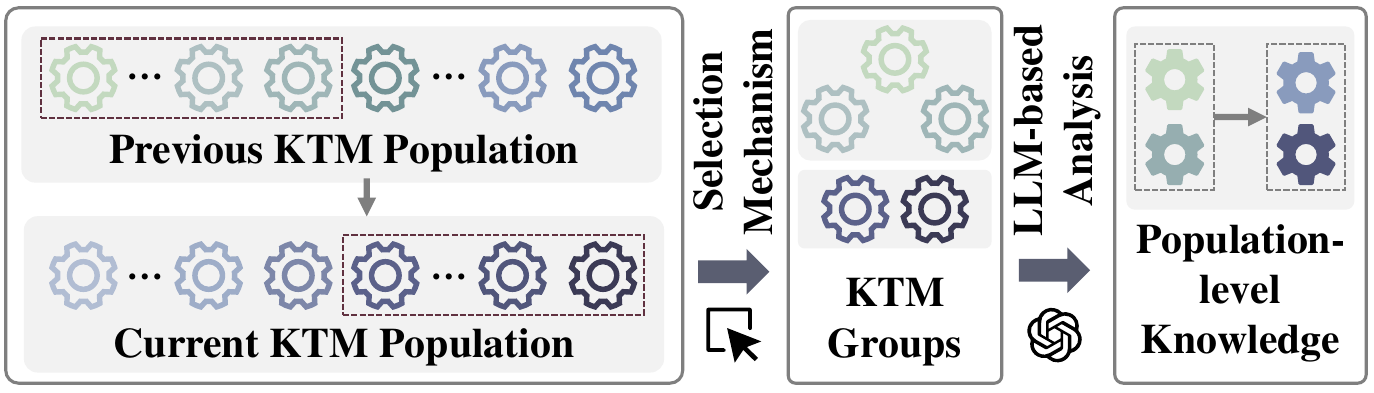}
	\caption{Illustration of the population-level knowledge acquisition mechanism.}
	\label{fig:gka}
  \end{figure}
The population-level knowledge acquisition mechanism is designed to capture coarse-grained evolutionary regularities across successive KTM populations. 
As illustrated in Fig. \ref{fig:gka}, this process begins by selecting high-performing KTMs from the top half of the current population and underperforming KTMs from the bottom half of the previous generation, with selection governed by the following probability distributions:
\begin{equation}
p_{i}^{h} = \frac{(1 - F_i' + \varepsilon)}{\sum_{i=1}^{N/2} (1 - F_i' + \varepsilon)} 
\qquad
p_{i}^{u} = \frac{(F_i' + \varepsilon)}{\sum_{i=1}^{N/2} (F_i' + \varepsilon)}
\label{pro}
\end{equation}
where $p_{i}^{h}$ and $p_{i}^{u}$ represent the probabilities of selecting the $i$-th high-performing and underperforming KTM, respectively; $N$ is the size of the KTM population; $F_i':=\min(F_i,1)$ is the clipped performance value; $F_i$ denotes the performance of the $i$-th KTM as defined in Eq. \ref{eq:fitness}, where smaller values of $F_i$ correspond to better performance; and $\varepsilon$ is a small constant introduced for numerical stability.
Based on the above probabilities, $N_{sel}$ KTMs are selected from each subset to form their respective groups, where $N_{sel}$ is sampled as a random integer satisfying $2\leq N_{sel} \leq \left\lceil  N /4  \right\rceil$.
The two groups are analyzed by the LLM using the prompt illustrated in Fig. \ref{prompt:GKA}, which guides the LLM to synthesize concise knowledge for future KTM refinement.
The population-level knowledge acquisition mechanism is triggered when the population changes, specifically when a superior KTM replaces an existing one.

The population-level knowledge acquisition mechanism seeks to uncover broader evolutionary trends that characterize the overall search trajectory, while the individual-level knowledge acquisition mechanism focuses on fine-grained transformations between individual KTMs.
In contrast to individual-level improvement knowledge, which is specific to individual transformations, the population-level counterpart derives generalized principles that capture emergent regularities across the search space.
By integrating both individual- and population-level knowledge, the SKTD algorithm achieves a synergistic balance between localized precision and strategic foresight, enabling effective exploration of the KTM landscape.

\begin{figure}[!t]
	\centering
	\fcolorbox{black}{gray!10}{\parbox{0.9\linewidth}{
	\begin{center}
	\textcolor[rgb]{0.0,0.0,0.0}{\textbf{Population-level Knowledge Acquisition}}\\
	\end{center}
	\textcolor[rgb]{0.7, 0.3, 0.0}{\textbf{\textit{Description}:}
	You are given two populations of Python functions named `LLMTransfer', each representing a different category of transfer optimization methods. The first population consists of underperforming methods, while the second population demonstrates superior performance.\\
	Your task is to conduct a step-by-step contrastive analysis between these two populations. Identify and explain the key differences in algorithmic design and transfer mechanisms. Then, synthesize actionable insights or improvement strategies for enhanced performance of the `LLMTransfer'.\\
	The underperforming LLMTransfer methods:\\
	\#UDKTM\#\\
	The high-performing LLMTransfer methods:\\
	\#HPKTM\#\\
	Finally, summarize your ideas / thoughts / suggestions in a single graph and within 200 words using the following format:
	\#KLG-XML-FORMAT\#\\
	}
	}}
	\captionof{figure}{Prompt template for population-level knowledge acquisition}
	\label{prompt:GKA}
  \end{figure}

\subsubsection{Knowledge-Guided KTM Improvement}

The acquired individual- and population-level knowledge entries are stored in a unified knowledge base \(\mathcal{K} = \{(K_i, \mathcal{L}_i)\}_{i=1}^n\) to support the \textit{improve} operator, where each \(K_i\) represents a textual improvement insight and \(\mathcal{L}_i\) denotes the set of underperforming KTMs from which this knowledge was extracted.
To maintain the integrity and diversity of \(\mathcal{K}\), a semantic de-duplication mechanism is applied.
Specifically, each knowledge entry is embedded into a high-dimensional vector space upon generation \cite{nie2024text}. The semantic similarity between the new embedding and existing embeddings is measured using cosine similarity \cite{vijaymeena2016survey}, and the new entry is discarded if the similarity exceeds a predefined threshold to avoid redundant information in \(\mathcal{K}\).
Following the de-duplication process, the filtered knowledge is organized into \(\mathcal{K}\).

Once \(\mathcal{K}\) starts accumulating knowledge, the \textit{improve} operator is activated by first selecting an original KTM, denoted as \(\text{KTM}^{ori}\), according to the probability $p_i^h$ defined in Eq. \ref{pro}.
Subsequently, the selected \(\text{KTM}^{ori}\) is processed by Alg. \ref{alg:improve} to produce an improved KTM, denoted as \(\text{KTM}^{imp}\).
This procedure begins by embedding \(\text{KTM}^{ori}\), denoted as \(e_{ktm}\) (line 1). 
Subsequently, for each item \((K_i, \mathcal{L}_i) \in \mathcal{K}\), a composite string \(t_i\) is constructed by concatenating \(K_i\) with the raw textual form of \(\mathcal{L}_i\) (line 3), which is then embedded and denoted as \(e_i\) (line 4).
The similarity score \(s_i\) is computed by evaluating the cosine similarity between \(e_{ktm}\) and each \(e_i\) (line 5). 
Based on \(\{s_i\}\), the most relevant knowledge entries are identified and aggregated to form the knowledge set \(\mathcal{S}\) (line 7). 
A prompt is subsequently constructed based on \(\mathcal{S}\), as shown in Fig. \ref{prompt:improvero}, to guide the LLM to generate \(\text{KTM}^{imp}\) (line 8).
In this prompt, the placeholders \#KTM-ORI\#, \#KNOWLEDGE\#, and \#IMP-XML-FORMAT\# correspond respectively to \(\text{KTM}^{ori}\), \(\mathcal{S}\), and the required output format.

\begin{algorithm}[htbp]
	\caption{Pseudocode of the \textit{improve} operator.}
	\SetArgSty{textnormal}
	\label{alg:improve}
	\KwIn{
		\\\quad \(\text{KTM}^{ori}\): Selected KTM for \textit{improve} operation. \\
		\\\quad \(\mathcal{K} = \{(K_i, \mathcal{L}_i)\}\): Knowledge base. \\
	}
	\KwOut{
		\\\quad \(\text{KTM}^{imp}\): Improved KTM.
	}
	\(e_{ktm} \leftarrow \text{Embed}(\text{KTM}^{ori})\) \\
	\ForEach{\((K_i, \mathcal{L}_i) \in \mathcal{K}\)}{
		\(t_i \leftarrow \text{Concatenate}(K_i, \text{Text}(\mathcal{L}_i))\) \\
		\(e_i \leftarrow \text{Embed}(t_i)\) \\
		\(s_i \leftarrow \text{Similarity}(e_{ktm}, e_i)\) \\
	}
	\(\mathcal{S} \leftarrow \text{Top}(K_i \text{ ranked by } s_i)\) \\
	\(\text{KTM}^{imp} \leftarrow \text{LLM}(\text{Prompt}(\text{KTM}^{ori}, \mathcal{S}, \textit{improve}))\) \\
\end{algorithm}
\FloatBarrier

\begin{figure}[htbp]
	\centering
	\fcolorbox{black}{gray!10}{\parbox{0.9\linewidth}{
	\begin{center}
	\textcolor[rgb]{0.0,0.0,0.0}{\textbf{Improve}}\\
	\end{center}
	\textcolor[rgb]{0.9,0.3,0.1}{\textbf{\textit{Description}:}
	Given a Python function named LLMTransfer which implements an evolutionary transfer optimization method, your task is to \textbf{design an improved version of this function by applying the provided performance enhancement tips or ideas}, while ensuring that the function name remains LLMTransfer, the input/output format remains unchanged, and the core functionality is preserved but enhanced for better performance. Here is the Python code of LLMTransfer function which you need to improve and its format you need obey:\\
	Original LLMTransfer function:
	\#KTM-ORI\#\\
	Function format specification:
	\#FORMAT\#\\
	Improvement tips or ideas:
	\#KNOWLEDGE\#\\
	Provide the newly designed LLMTransfer function using the following XML format:
	\#IMP-XML-FORMAT\#\\
	}
	}}
	\captionof{figure}{Prompt template for the \textit{improve} operator}
	\label{prompt:improvero}
\end{figure}
\FloatBarrier

\subsection{KTM Repair and Refinement Module} \label{subsec:generation}

Across the preceding components, the operators \textit{initialize}, \textit{recombine}, \textit{discover}, and \textit{improve} all rely on a shared KTM repair and refinement module to produce new KTMs, which is elaborated in this subsection.
This module is triggered by operator-specific prompts that instruct the LLM to generate the corresponding KTM, as illustrated in the left part of Fig. \ref{fig:workflow}.
Upon receiving the prompt for a specific operator, the LLM generates a response from which the KTM code is extracted. 
The extracted code is first checked to ensure executability and compliance with the required input-output format, after which the validated KTM is returned to the corresponding search process.
When issues are detected in the LLM's response or the extracted code, an iterative correction process is triggered.
Specifically, code execution errors are addressed through a code repair mechanism guided by error feedback, in which both the prior interaction messages and the execution error information are provided to the LLM to regenerate an improved response \cite{huang2025autonomous}. Otherwise, or after unsuccessful retries, the original prompt is reused to regenerate the KTM until a valid implementation is obtained.
As a result, this module ensures that all generated KTMs are executable and format-compliant.

\section{Experimental Study} \label{sec:experiment}
In this section, we conduct comprehensive empirical studies to evaluate the effectiveness of our proposed LLM-assisted knowledge transfer design method across EMTO scenarios with varying levels of task similarity.

\subsection{Experimental Configurations} \label{subsec:exp_set}
To assess the effectiveness of the proposed method, we construct nine multi-task optimization scenarios using a recently proposed scalable test problem generator \cite{generator}.
This generator enables flexible scenario construction under specified configurations, ensuring that the generated tasks satisfy the preset settings and thereby facilitating a more thorough investigation of the KTM behavior across different conditions.
The detailed generation rules are summarized in table \ref{tab:setting}, where the columns ``NUMT'' and ``DIM'' denote the number of optimization tasks in each scenario and the dimensionality of the decision variables for each task, respectively. 
Specifically, a base task is adopted to generate scenarios under different transfer scenarios (TSs) and similarity distributions (SDs), with all configuration settings taken directly from \cite{generator}.
When the transfer scenario is $T_a$, all tasks within a scenario are derived from the same task family, whereas under $T_e$, they come from different families.
To ensure the reliability of the experimental results, ten representative task families are adopted, including Sphere, Ellipsoid, Schwefel 2.2, Quartic with noise, Ackley, Rastrigin, Griewank, Levy, Rosenbrock, and Weierstrass, with their respective test functions defined in \cite{generator,Benchmark}.

\begin{table}[htbp]
	\centering
	\caption{Configurations of the test scenarios.}
	\begin{tabular}{cccccccc}
		\toprule
		Benchmarks&Similarity&NUMT&DIM  & TS & SD \\
		\midrule
		P1&&2&50&$T_e$&$h_1^{h}$ \\
		P2&High&5&50&$T_a$&$h_2^{h}$ \\
		P3&&10&50&$T_e$&$h_2^{h}$ \\
		\midrule
		P4&&2&50&$T_a$&$h_1^{m}$ \\
		P5&Medium&5&50&$T_e$&$h_2^{m}$ \\
		P6&&10&50&$T_e$&$h_3^{m}$ \\
		\midrule
		P7&&2&50&$T_a$&$h_2^{l}$ \\
		P8&Low&5&50&$T_e$&$h_1^{l}$ \\
		P9&&10&50&$T_e$&$h_2^{l}$ \\
		\bottomrule
	\end{tabular}
	\label{tab:setting}
\end{table}

To evaluate the performance of KTMs generated by the proposed method, we conduct a comparative analysis against several established knowledge transfer approaches. Specifically, four well-known hand-crafted methods are selected: vertical genetic crossover-based transfer (namely VC) \cite{VC}, orthogonal transfer (namely OT) \cite{OT}, autoencoding-based solution mapping (namely AE) \cite{AE}, and affine transformation (namely AF) \cite{AF}. Moreover, we also include a recently introduced LLM-based program search method, denoted as EoH \cite{EoH}, which is employed to automatically generate KTMs for comprehensive comparisons.
The performance of a given KTM $m$ under a given scenario is quantified by the normalized fitness value, defined as
\begin{equation}
	\label{eq:fitness}
	F_{m} = \frac{1}{L} \frac{1}{N_t} \sum_{l=1}^L \sum_{i=1}^{N_t} (f(i,l) / f_{single}(i))
\end{equation}
where $N_t$ is the number of tasks; $L$ is the number of independent runs to assess the selected KTM, and is set to 20 in this study; $f(i,l)$ is the objective value of task-$i$ in the $l$-th run; and $f_{single}(i)$ is the mean objective value obtained by applying the basic optimizer on task-$i$ in isolation across $L$ runs.
Note that $F_{m} \geq 0$, and a smaller value indicates superior optimization performance in this study.
For each optimization task, the basic optimizer is configured as a genetic algorithm implemented using the pymoo framework \cite{pymoo}.
The basic optimizer equipped with different KTMs is configured with a population size of 100 and a maximum of 100 generations, resulting in a total of 10,000 fitness evaluations per run.
For all KTMs, according to \cite{AE}, the interval of knowledge transfer across tasks is set to 10, and the number of transferred solutions is capped at 10.

In the SKTD algorithm, based on the normalized fitness metric defined in Eq. \ref{eq:fitness}, two KTMs are considered to exhibit similar performance if the absolute difference between their fitness values is less than 0.01. This threshold is adopted as the criterion for the KTM de-duplication mechanism introduced earlier \cite{kalousis2004data}.
In addition, the \textit{improve} operator is applied with a probability of 0.2. When this operator is not applied, the \textit{recombine} and \textit{discover} operators are selected with probabilities of 0.7 and 0.3, respectively.
A higher probability is assigned to the \textit{recombine} operator to promote the exploitation of existing KTMs in generating promising ones, while reducing potential adverse effects caused by excessive exploration and premature convergence \cite{huang2025autonomous}.
In the self-guided knowledge extraction mechanism, the de-duplication threshold for filtering acquired knowledge is set to 0.85, and the \textit{improve} operator uses the top 10 most relevant knowledge entries, thereby retaining an adequate amount of informative and distinctive knowledge \cite{fernando2024promptbreeder}.
For the LLM-based algorithms, the detailed settings are configured based on \cite{huang2025autonomous,EoH} and summarized as follows:

\begin{itemize}[nosep]
	\item Number of independent trial runs: 3.
	\item Size of the KTM population: 10.
	\item Maximum number of generations: 10.
	\item Language model used for KTM generation: qwen-max \footnote{The LLM \textit{qwen-max} and \textit{text-embedding-v4} are developed by Alibaba and accessible via the Bailian platform. \url{https://bailian.console.aliyun.com/}.} (temperature: 1; maximum token limit: 8192).
	\item Text embedding model: text-embedding-v4 with 1536 dimensions.
\end{itemize}

\subsection{Results and Discussion}

\subsubsection{Performance Comparison with Baseline Algorithms}

The comparative results of the normalized fitness values obtained by the proposed SKTD algorithm against all the other compared baseline algorithms across the test scenarios are presented in table \ref{tab:res_all}.
This table reports the mean performance of hand-crafted KTMs over 20 runs. 
For LLM-based methods, according to \cite{huang2025autonomous}, three independent trial runs are conducted. In each trial, all generated KTMs are evaluated over 20 runs, including the final best KTM selected from that trial. The normalized fitness values of the best KTMs obtained from the three trials are then averaged to produce the reported value.
Note that a reported value of 0.000 indicates that the result is smaller than 0.001 but greater than zero.
Compared with hand-crafted KTMs, SKTD exhibits clear advantages in both effectiveness and generality.
Specifically, the effectiveness of VC and AE degrades markedly as task similarity decreases.
In the high-similarity scenarios (P1-P3), both methods generally achieve normalized fitness values below 1, indicating beneficial knowledge transfer, whereas in the low-similarity scenarios (P7-P9), all reported normalized fitness values exceed 1, suggesting that the transferred knowledge becomes ineffective or even detrimental.
Moreover, severe negative transfer is observed in several cases, most notably VC on P8 and AE on P4, where the obtained normalized fitness values are more than 30\% worse than their basic single-task optimizer.
For AF and OT, a different performance trend can be observed.
Under comparable similarity levels, AF achieves better performance in the $T_a$ setting than in $T_e$, whereas OT shows stronger capability to handle the $T_e$ scenarios, particularly in the medium- and low-similarity cases.
OT consistently achieves higher normalized fitness on P4 and P7, exceeding its corresponding values on P5-P6 and P8-P9 by more than 0.1, respectively.
In contrast, AF exhibits an opposite trend, with its normalized fitness on P4 and P7 being over 0.1 lower than those observed on the corresponding comparison problems.
These observations indicate that hand-crafted KTMs tend to rely on specific scenario conditions to achieve competitive performance.
In contrast, SKTD achieves normalized fitness values below 0.7 across all test scenarios, demonstrating its ability to autonomously construct effective KTMs under varying conditions.

\begin{table}[htbp]
	\centering
	\caption{Normalized fitness values obtained by different knowledge transfer methods over 20 independent runs. Superior results for each scenario are highlighted in bold font.}
	\begin{tabular}{ccccccc}
		\toprule
		Benchmarks & VC & OT & AE & AF & EoH & SKTD \\
		\midrule
		P1&0.563&0.302&0.502&0.752&\textbf{0.000}&0.004\\
		P2&1.036&\textbf{0.338}&0.992&0.493&0.369&0.355\\
		P3&0.821&0.773&0.927&0.757&\textbf{0.653}&0.673\\
		\midrule
		P4&1.094&1.083&1.359&0.798&\textbf{0.562}&0.645\\
		P5&1.068&0.788&1.070&0.918&0.630&\textbf{0.606}\\
		P6&0.940&0.803&1.023&0.897&0.699&\textbf{0.581}\\
		\midrule
		P7&1.189&0.931&1.186&0.721&0.742&\textbf{0.659}\\
		P8&1.300&0.813&1.181&1.017&0.682&\textbf{0.553}\\
		P9&1.075&0.828&1.192&1.010&0.921&\textbf{0.653}\\
		\bottomrule
	\end{tabular}
	\label{tab:res_all}
\end{table}

When comparing LLM-based approaches, SKTD consistently outperforms EoH in more test scenarios. 
Specifically, SKTD achieves the best performance among all tested algorithms in five out of the nine scenarios, whereas EoH attains the top results in only three cases.
This advantage becomes particularly pronounced in the low-similarity test scenarios, where SKTD achieves markedly lower normalized fitness values.
In P9, the normalized fitness value obtained by EoH is above 0.9, suggesting limited effectiveness of the EoH-generated KTMs under certain low-similarity scenarios.
The observed performance difference is likely related to the fundamental distinction in the underlying mechanisms of the two methods. Although both methods can be broadly categorized as LLM-based search approaches, their search processes differ significantly.
General autonomous programming frameworks typically focus on improving existing solvers by using LLM-based operators to generate new solver variants. EoH, for instance, first represents a heuristic as a natural-language thought and then translates it into executable code, thereby using the thought stage to guide solver generation and refinement.
Although this mechanism enhances the generation of new solver variants, it still provides limited explicit guidance on where the search should focus. Consequently, its performance may be limited in more challenging EMTO scenarios, where effective search direction becomes increasingly important.
By contrast, SKTD extracts structured knowledge from the KTM evolution process and incorporates it into subsequent KTM generation, thereby guiding later search toward more effective KTM designs.
Notably, in the low-similarity scenarios, hand-crafted KTMs struggle to achieve satisfactory optimization performance, with all tested hand-crafted methods yielding normalized fitness values above 0.7, underscoring the need for autonomously designing effective KTMs adapted to such conditions.
However, general autonomous programming may exhibit limited effectiveness under such challenging conditions, whereas SKTD achieves more robust performance.

\subsubsection{Analysis of Search Dynamics in SKTD}

\begin{figure*}[htbp]
	\centering
	\subfigure[\scriptsize P1]{
	\includegraphics[width=0.23\columnwidth]{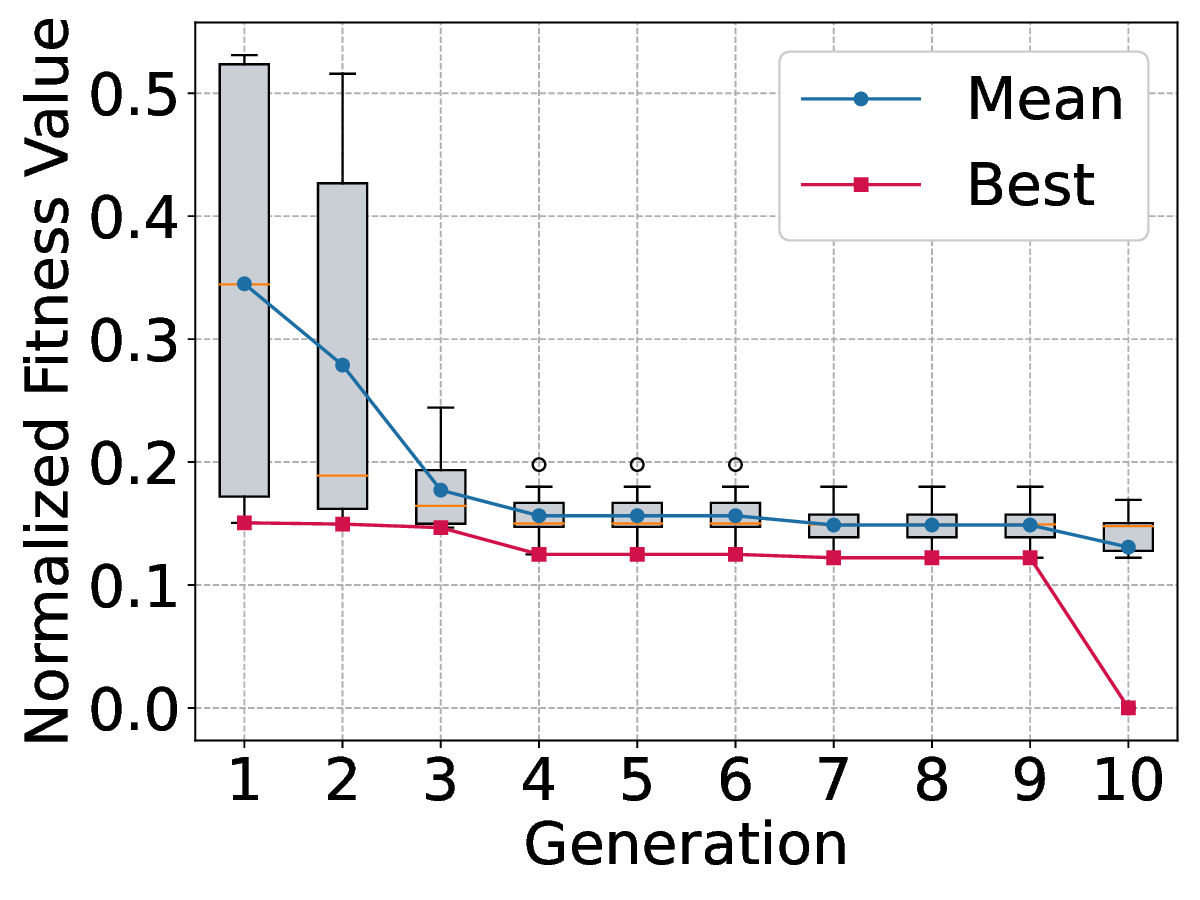}}
	\subfigure[\scriptsize P21]{
	\includegraphics[width=0.23\columnwidth]{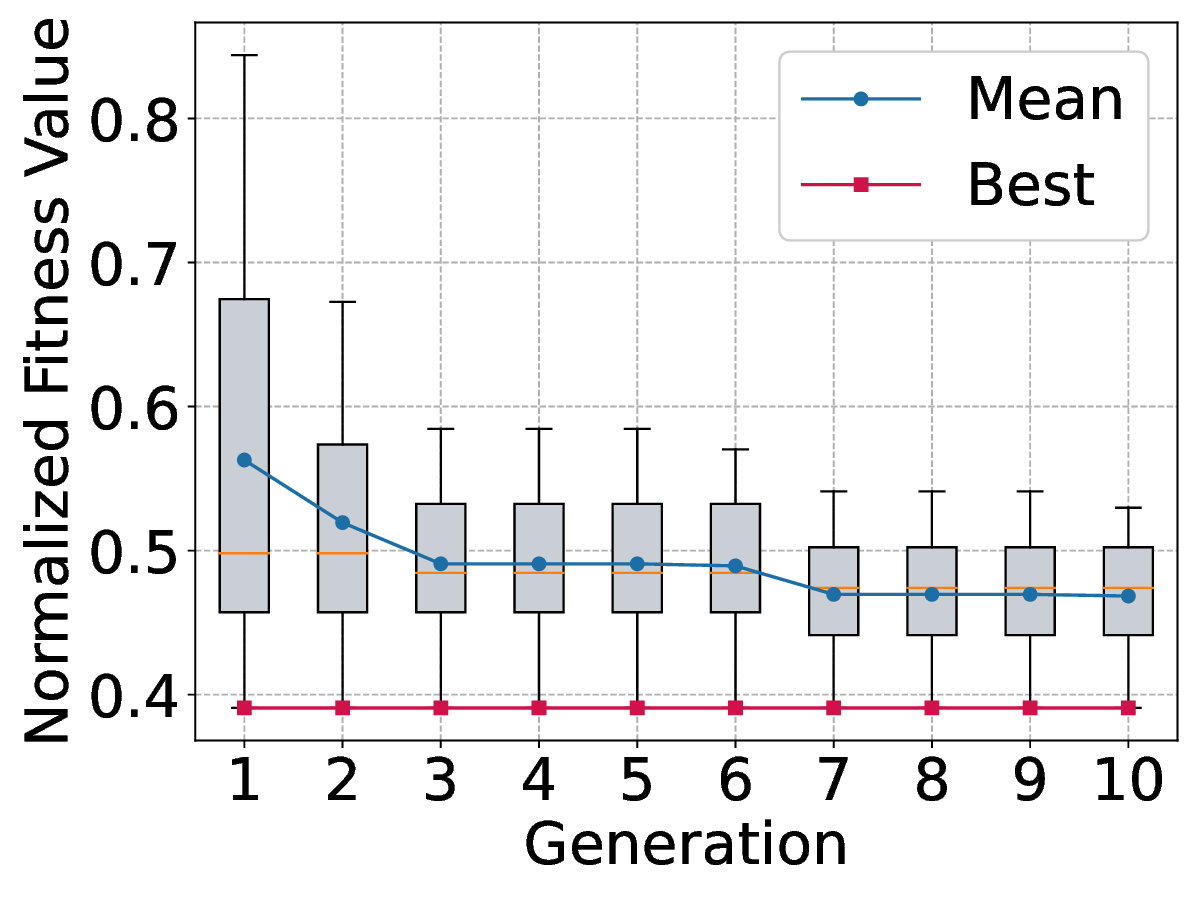}}
	\subfigure[\scriptsize P5]{
	\includegraphics[width=0.23\columnwidth]{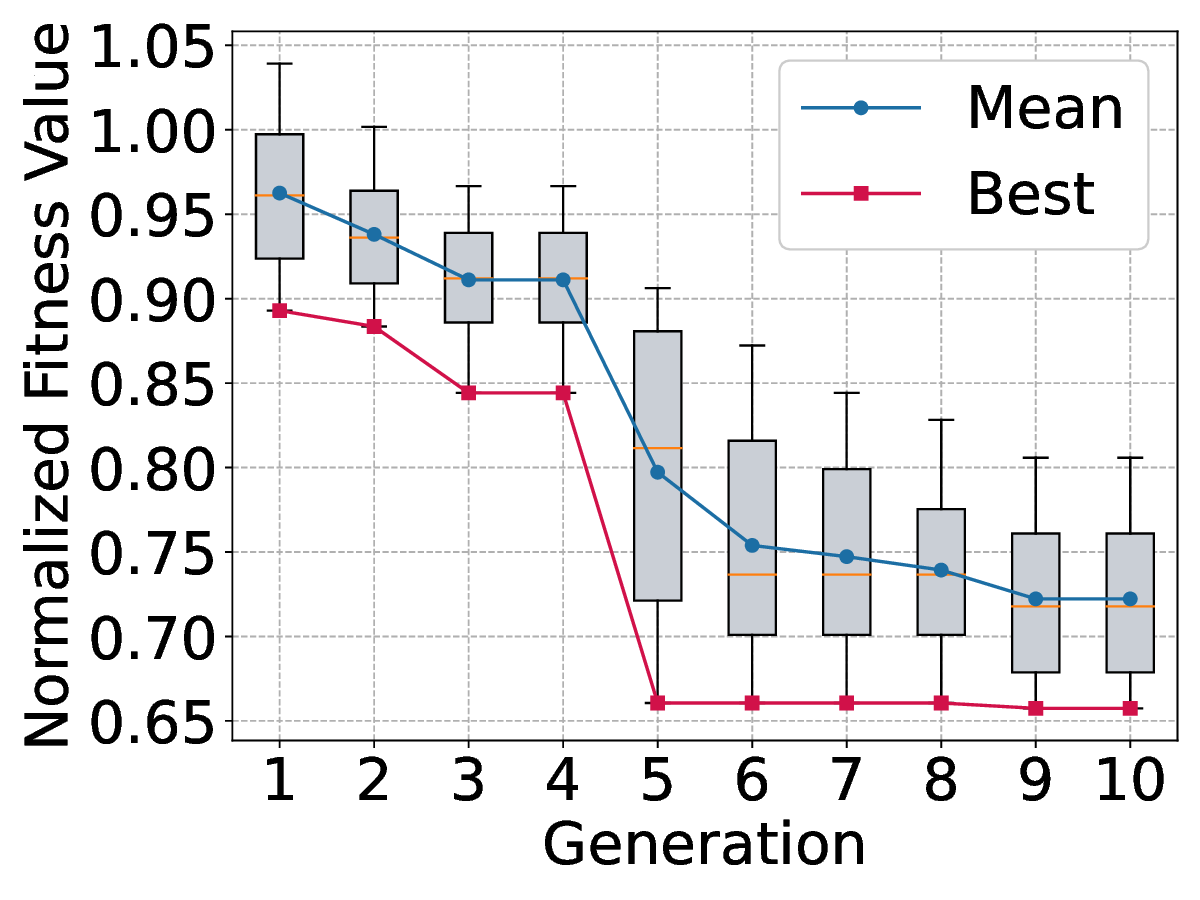}}
	\subfigure[\scriptsize P8]{
	\includegraphics[width=0.23\columnwidth]{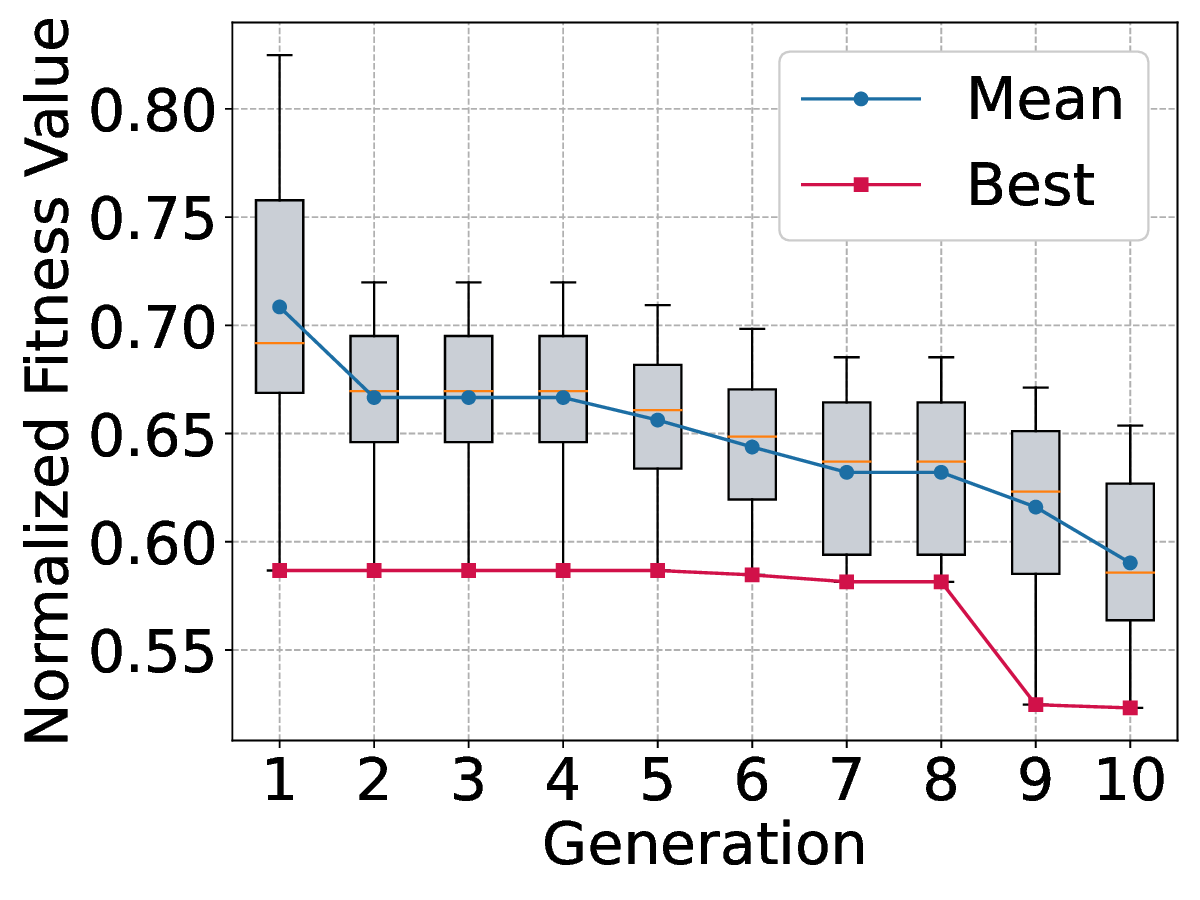}}
	\caption{Convergence curves of normalized fitness values achieved by the proposed SKTD algorithm on representative instances. Y-axis: normalized fitness value; X-axis: generations.}
	\label{fig:box}
\end{figure*}

To further analyze the search process of KTMs within SKTD, Fig. \ref{fig:box} presents boxplots of the normalized fitness values of KTMs across generations for representative benchmark-run instances, which are consistently used in subsequent analyses to ensure experimental rigor.
As observed in Fig. \ref{fig:box}, the mean and best normalized fitness values of KTMs on the representative instances exhibit a noticeable decreasing trend across generations.
Although the improvement of the best KTM across generations is sometimes marginal, the KTMs in the population still demonstrate steady progress, and the best KTM in the population may achieve further enhancements in the subsequent search process.
As shown in Fig. \ref{fig:box}(d), the best KTM shows no apparent improvement during generations 1-5, but exhibits noticeable improvements in generation 9.
Nevertheless, despite the lack of improvement in the best KTM, the overall KTM population exhibits clear progress, with noticeable decreases in the median, mean, and first-quartile normalized fitness values in generations 2 and 5.
These observed performance improvements of KTMs across generations demonstrate the effectiveness of the proposed SKTD algorithm in autonomously developing powerful KTMs.
Furthermore, the height of the box, defined as the interquartile range, reflects the diversity of the KTM population and is generally well preserved during optimization.
The interquartile range typically remains above 0.05, as shown in Fig. \ref{fig:box}(b)-(d).
This population diversity is supported by the KTM de-duplication mechanism, which removes KTMs with highly similar core mechanisms, thereby promoting the exploration of novel KTMs.

\begin{figure*}[!b]
	\centering
	\subfigure[\scriptsize P1]{
	\includegraphics[width=0.23\columnwidth]{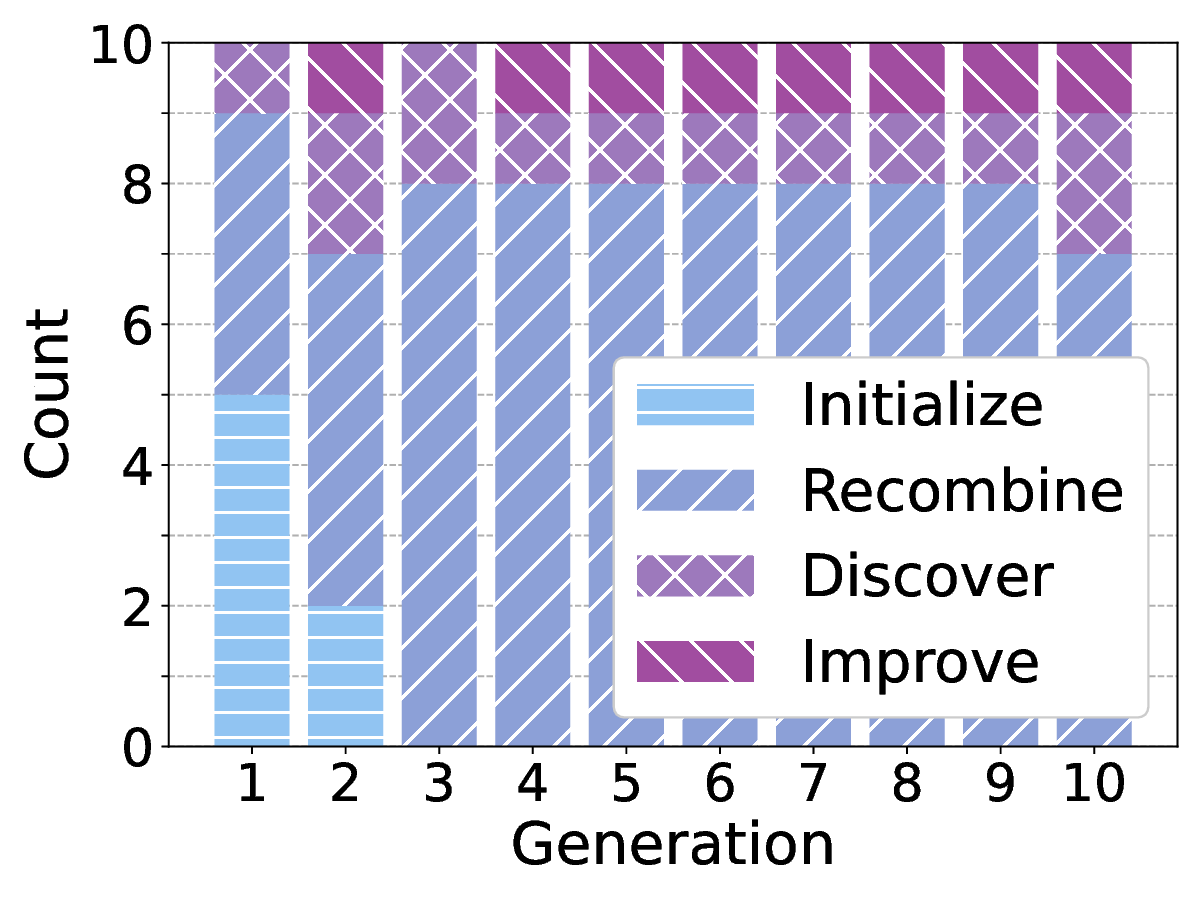}}
	\subfigure[\scriptsize P2]{
	\includegraphics[width=0.23\columnwidth]{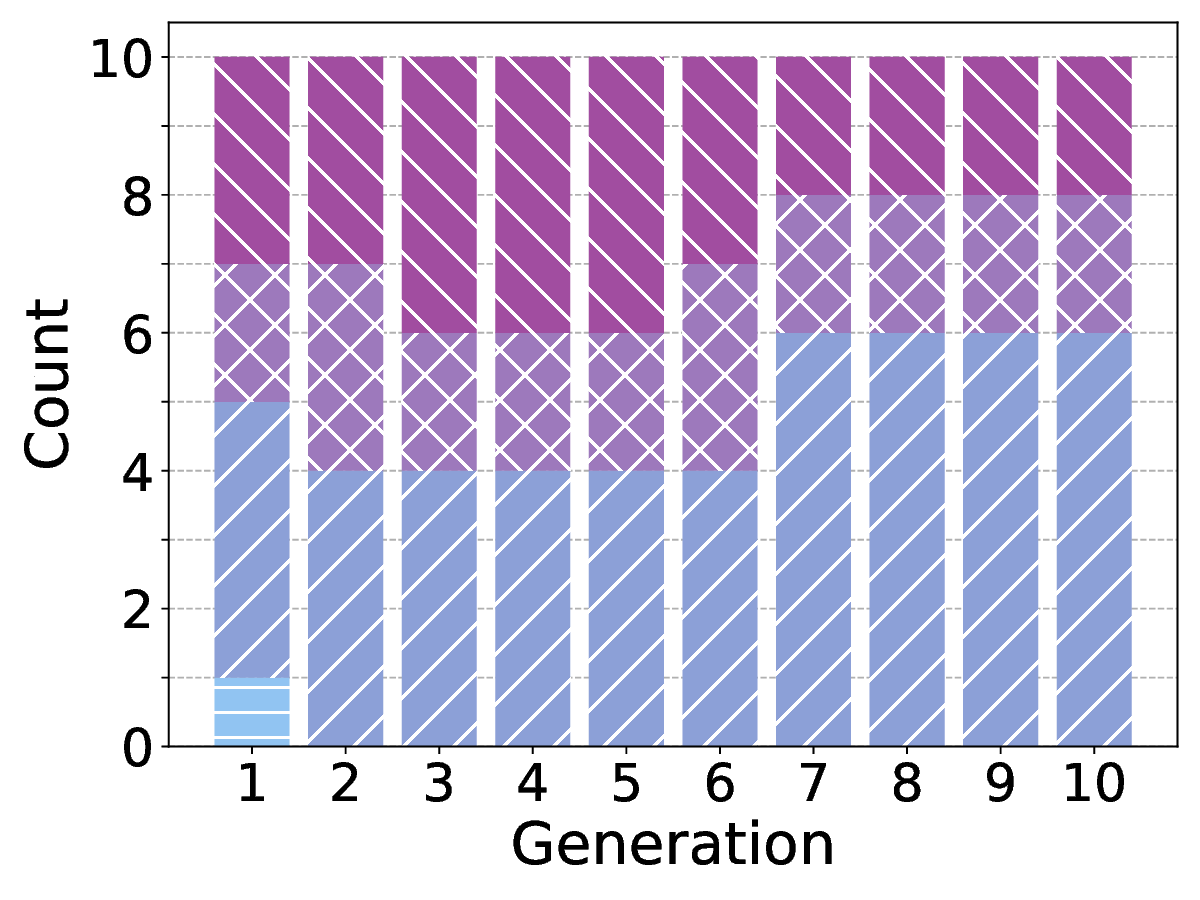}}
	\subfigure[\scriptsize P5]{
	\includegraphics[width=0.23\columnwidth]{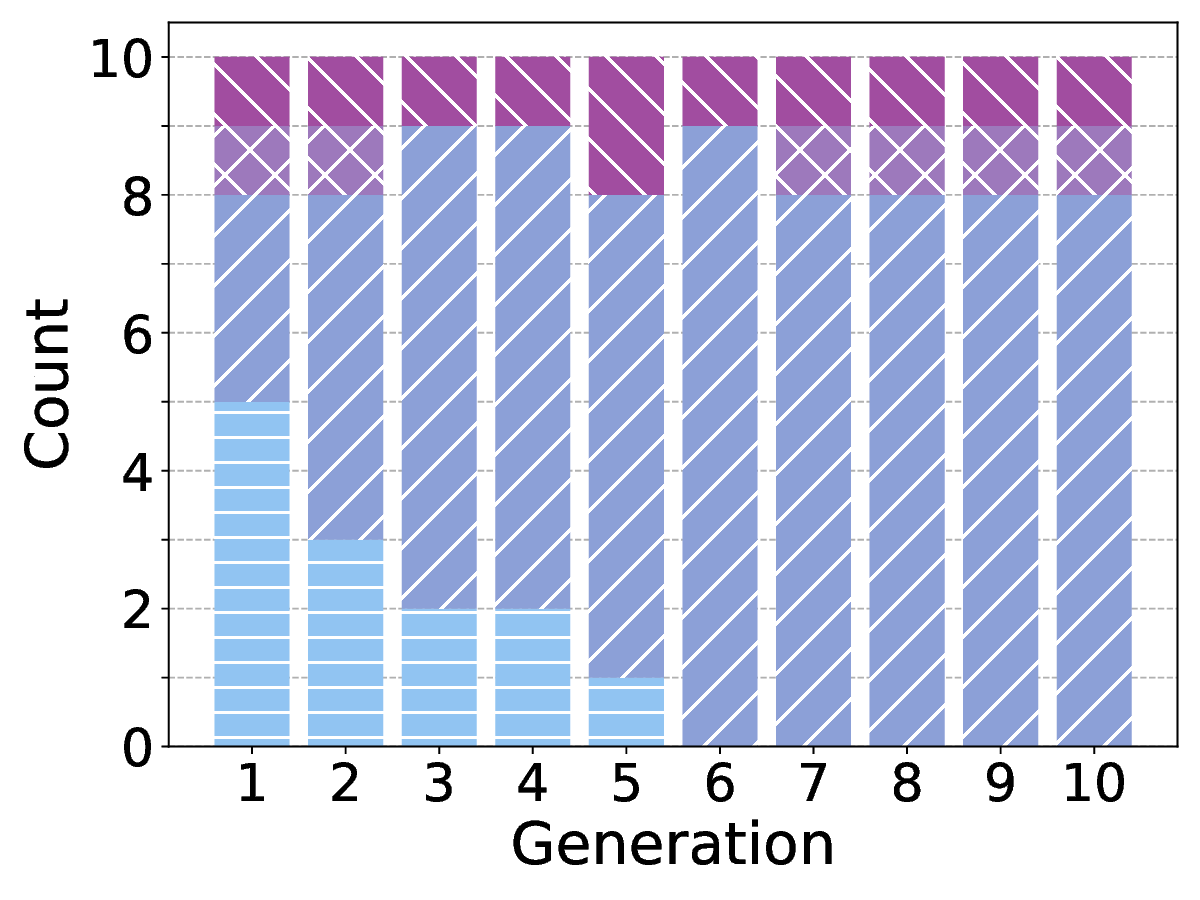}}
	\subfigure[\scriptsize P8]{
	\includegraphics[width=0.23\columnwidth]{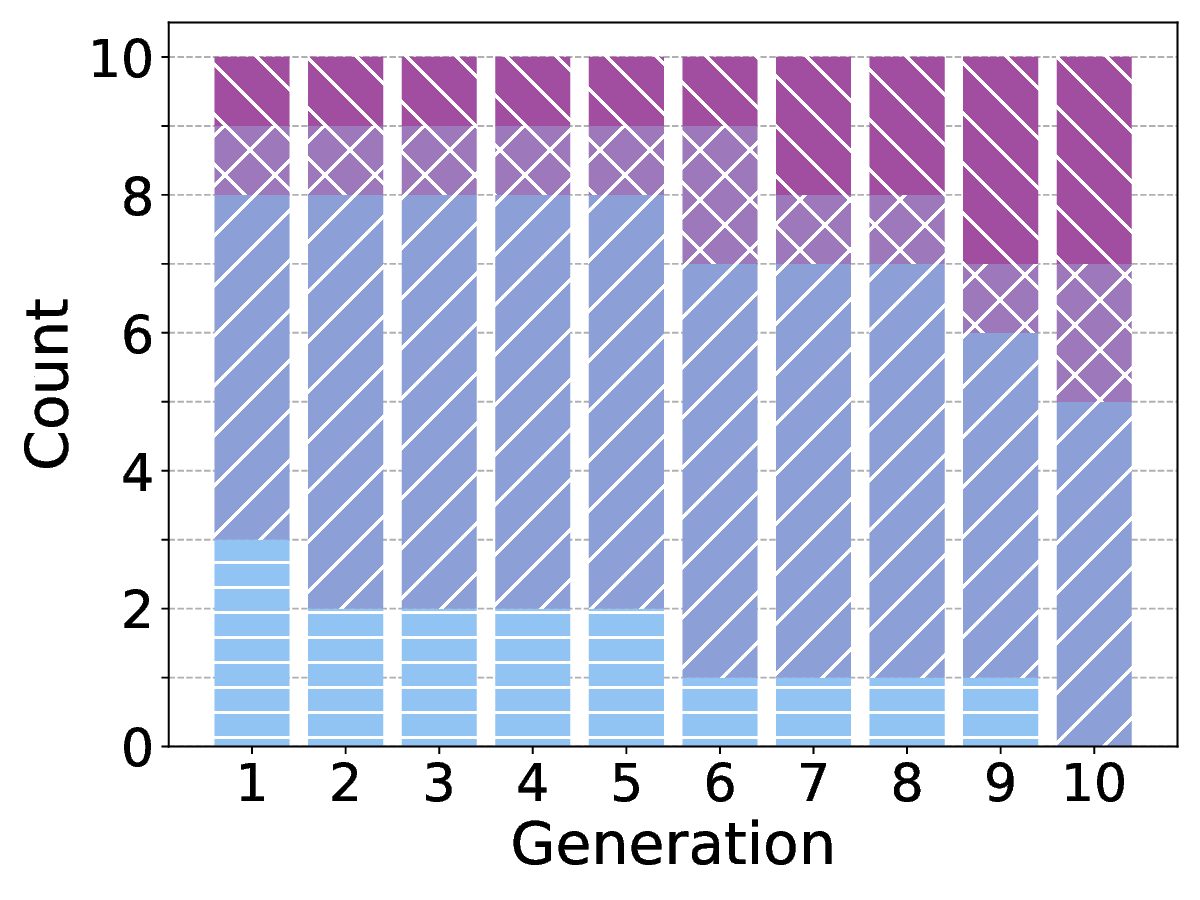}}
	\caption{Operators responsible for direct generation of current-population KTMs at different search stages of the proposed SKTD algorithm on representative instances. Y-axis: number of current-population KTMs directly generated by different operators; X-axis: generations.}
	\label{fig:op}
\end{figure*}
To gain deeper insights into the proposed operators, Fig. \ref{fig:op} illustrates how their contribution to KTM generation varies across the search process.
This figure depicts the distribution of current-population KTMs directly generated by different operators at each generation.
As shown in this figure, a notable proportion of KTMs is generated by the \textit{initialize} operator during the early stage of the search process.
However, as the search process progresses, the number of KTMs generated by the \textit{initialize} operator gradually decreases and eventually vanishes.
Moreover, by the end of the first generation, the fraction of KTMs originating from the \textit{initialize} operator has already fallen to half or less, further highlighting the effectiveness of the search dynamics.
The consistent emergence of KTMs generated by the \textit{recombine}, \textit{discover}, and \textit{improve} operators across generations further validates the effectiveness of these operators. The \textit{recombine} operator accounts for the largest proportion, providing strong evidence of its pivotal role in constructing high-quality KTMs.
Although the \textit{improve} operator is invoked less frequently than the \textit{discover} operator, its proportion occasionally surpasses that of the latter, suggesting its importance in generating high-performance KTMs, as shown in Fig. \ref{fig:op}(b) and (d).

\subsection{Insights of the Self-guided Knowledge Extraction Mechanism}

\subsubsection{Ablation Study}

\begin{table*}[!b]
	\centering
	\caption{Normalized fitness values of SKTD-NI and SKTD. Each method was run in three independent trials. The table reports trial-level results and their mean. For each benchmark, the better mean between the two methods is highlighted in bold, and among all trial-level results from both methods, the overall best one is also highlighted in bold.}
	\setlength{\tabcolsep}{7pt}
	\begin{tabular}{c|cccc|cccc}
		\toprule
		 \multirow{2}{*}{Benchmarks}&\multicolumn{4}{c|}{SKTD-NI}&\multicolumn{4}{c}{SKTD}\\
		 & R1& R2 & R3 & mean & R1 & R2& R3 &mean\\
		\bottomrule
		P1&\textbf{0.000}&\textbf{0.000}&\textbf{0.000}&\textbf{0.000}&\textbf{0.000}&\textbf{0.000}&0.011&0.004\\
		P2&\textbf{0.119}&0.431&0.379&\textbf{0.309}&0.391&0.410&0.263&0.355\\
		P3&0.544&0.748&0.718&\textbf{0.670}&\textbf{0.514}&0.774&0.731&0.673\\
		\midrule
		P4&0.736&0.845&0.790&0.790&0.563&0.816&\textbf{0.558}&\textbf{0.645}\\
		P5&0.808&0.689&\textbf{0.524}&0.674&0.657&0.525&0.634&\textbf{0.606}\\
		P6&0.595&0.609&0.566&0.590&\textbf{0.527}&0.615&0.600&\textbf{0.581}\\
		\midrule
		P7&0.628&0.781&0.758&0.723&\textbf{0.603}&0.695&0.680&\textbf{0.659}\\
		P8&0.526&0.567&0.561&\textbf{0.551}&\textbf{0.523}&0.543&0.593&0.553\\
		P9&0.605&0.732&0.649&0.662&\textbf{0.593}&0.675&0.690&\textbf{0.653}\\
		\bottomrule
	\end{tabular}
	\label{tab:ab_e}
\end{table*}

To further validate the effectiveness of the self-guided knowledge, an ablation experiment is conducted and the results are summarized in table \ref{tab:ab_e}. In this experiment, the proposed SKTD algorithm without the \textit{improve} operator is denoted as SKTD-NI. 
Table \ref{tab:ab_e} reports the normalized fitness values of the best KTM from each trial run, along with their mean values.
As can be observed, SKTD achieves higher mean best-KTM performance than SKTD-NI in a larger number of test scenarios.
Specifically, SKTD achieves better mean performance than SKTD-NI in P4-P7 and P9, demonstrating its effectiveness in the medium- and low-similarity test scenarios.
In these medium- and low-similarity scenarios (P4-P9) that demand the automatic generation of effective KTMs, SKTD-NI exhibits similarly poor performance to hand-crafted methods in several cases.
For example, in P4, SKTD-NI achieves 0.790, comparable to AF at 0.798; in P7, SKTD-NI attains 0.723, marginally worse than AF at 0.721, according to tables \ref{tab:res_all} and~\ref{tab:ab_e}.
By incorporating self-guided knowledge, SKTD achieves mean normalized fitness values below 0.7 across all test scenarios, demonstrating improved generality.
Moreover, according to the detailed results across the three trial runs, SKTD attains the best KTMs on seven test scenarios, whereas SKTD-NI does so on only three.
From a per-instance perspective, SKTD-NI yields normalized fitness values exceeding 0.7 in nine benchmark-run instances, whereas SKTD does so in only three.
These results demonstrate that incorporating the \textit{improve} operator not only strengthens the performance of the generated KTMs but also reduces the occurrence of poorly performing KTMs, thereby validating the effectiveness of the self-guided knowledge extraction mechanism.

\subsubsection{Insights of the \textit{improve} Operator}

\begin{figure*}[htbp]
	\centering
	\subfigure[\scriptsize P1]{
	\includegraphics[width=0.23\columnwidth]{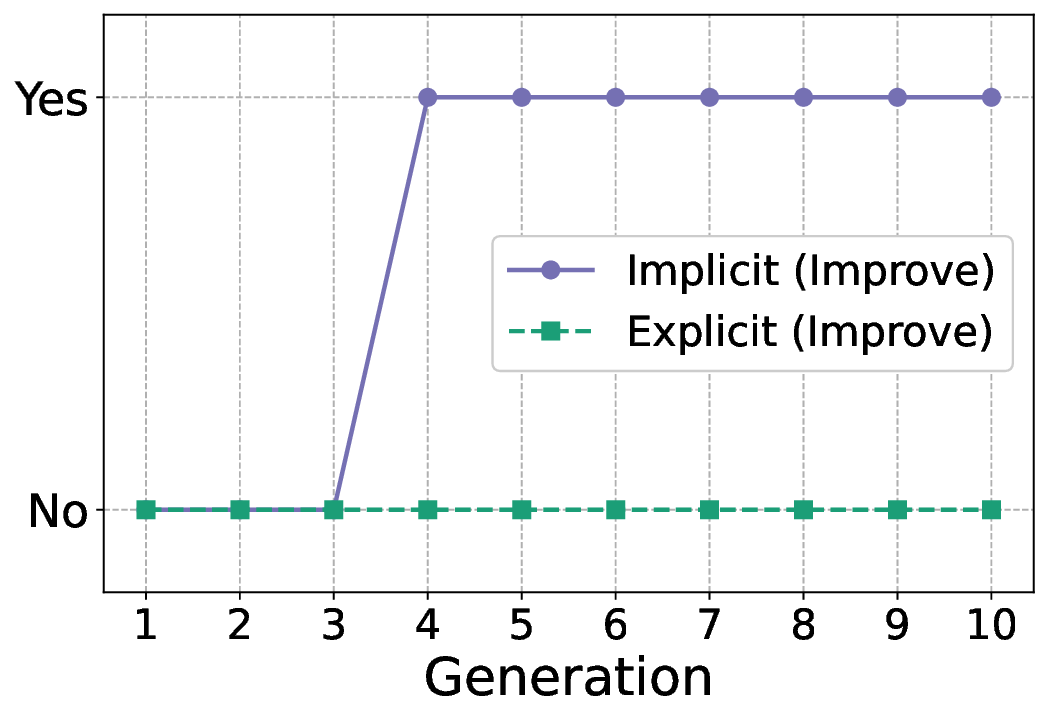}}
	\subfigure[\scriptsize P2]{
	\includegraphics[width=0.23\columnwidth]{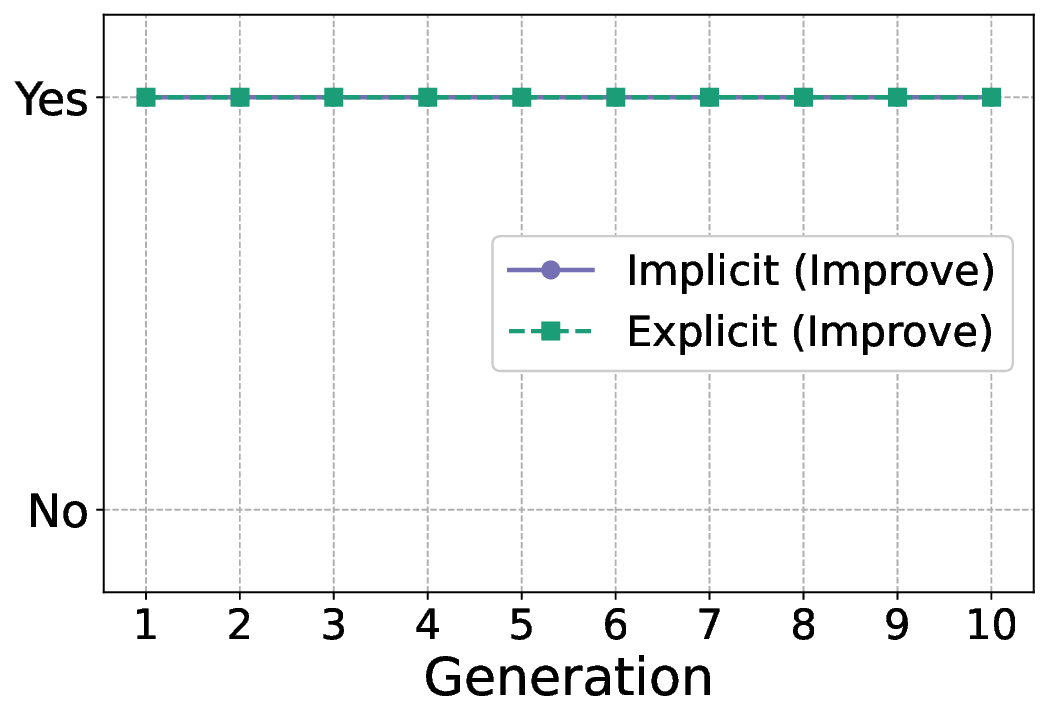}}
	\subfigure[\scriptsize P5]{
	\includegraphics[width=0.23\columnwidth]{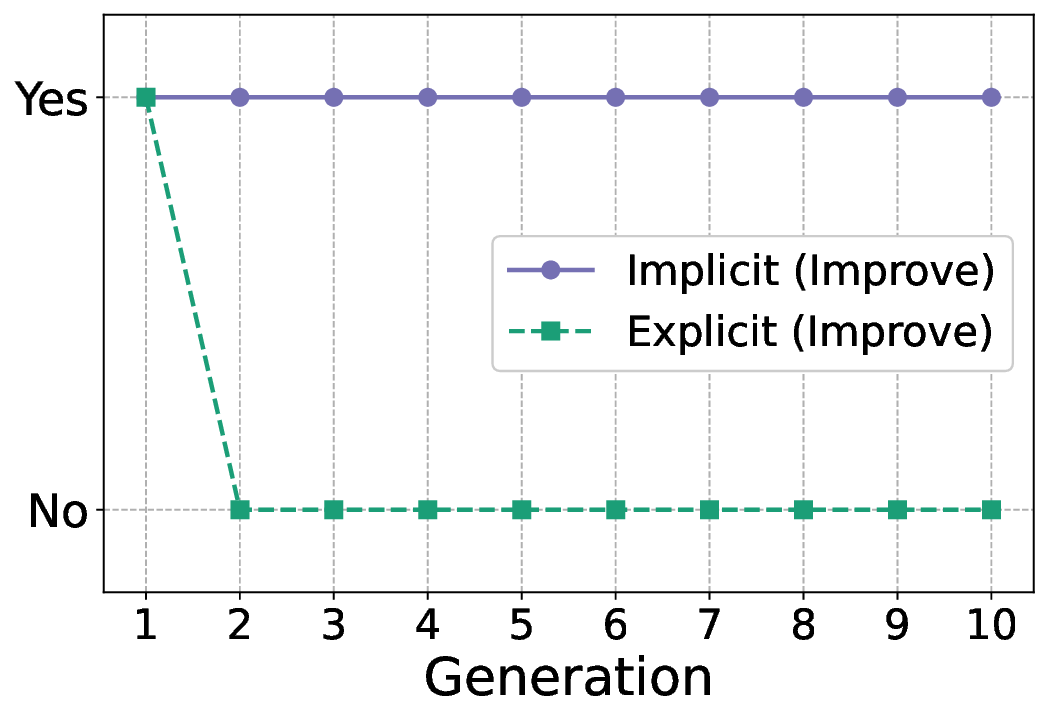}}
	\subfigure[\scriptsize P8]{
	\includegraphics[width=0.23\columnwidth]{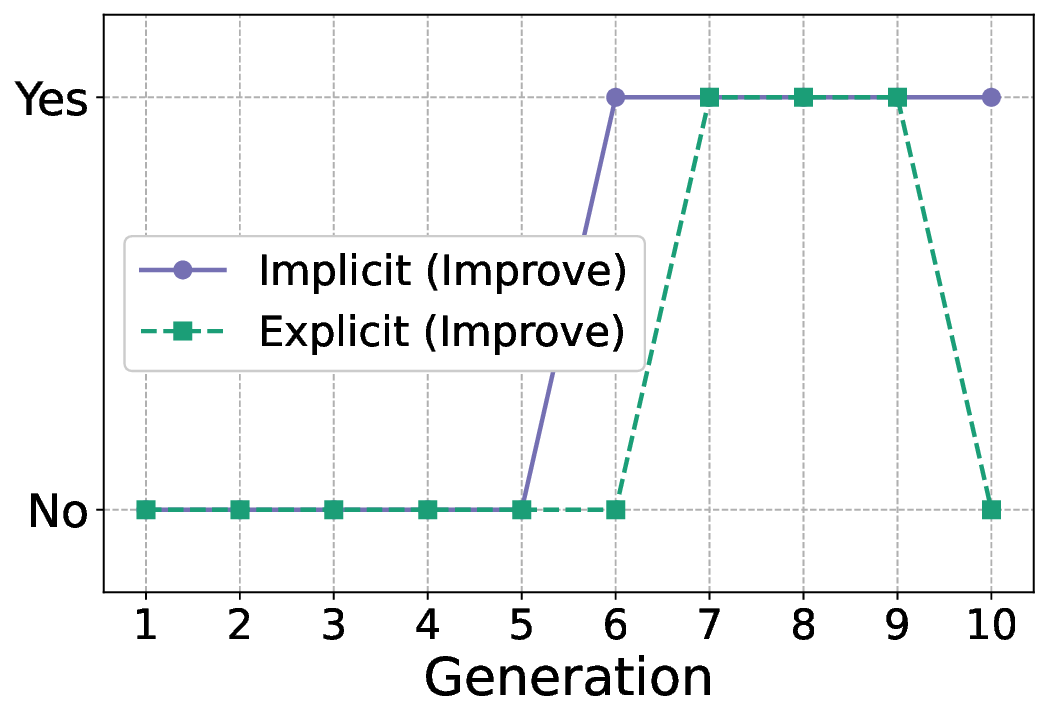}}
	\caption{The best KTM explicitly or implicitly generated by the \textit{improve} operation at each generation on representative instances. Y-axis: whether the best KTM is explicitly or implicitly generated by the \textit{improve} operation (Yes/No); X-axis: generations.}
	\label{fig:best_ktm}
\end{figure*}

To further investigate the contribution of the \textit{improve} operator, we examined the relationship between the best KTMs in the evolving population and this operator, as illustrated in Fig. \ref{fig:best_ktm}. 
In the figure, green dashed lines and purple solid lines indicate whether the best KTM in each generation is generated explicitly or implicitly by the \textit{improve} operator.
A KTM is regarded as explicitly produced by the \textit{improve} operator when it is directly generated by applying this operator to its parent KTMs. In contrast, a KTM is considered implicitly influenced by the \textit{improve} operator if at least one of its parents was directly generated through this operator.
As shown in Fig. \ref{fig:best_ktm}, the best KTMs are frequently associated with the \textit{improve} operator.
In certain cases, such as Fig. \ref{fig:best_ktm}(b), the best KTMs across the entire search process are generated explicitly by the \textit{improve} operator.
Although not all best KTMs are explicitly generated by the \textit{improve} operator, this operator typically injects self-guided knowledge into their parent KTMs, which is subsequently propagated to the offspring KTMs through other operators, as illustrated in Fig.~\ref{fig:best_ktm}(a) and (c).
Moreover, in the later stages of the search process, the best KTMs in the population are generally linked either explicitly or implicitly to the \textit{improve} operator, as illustrated in Fig.~\ref{fig:best_ktm}(a) and (d).
These observations suggest that high-performing KTMs are typically associated with the \textit{improve} operator, thereby reinforcing the effectiveness of the proposed self-guided knowledge extraction mechanism.

\begin{figure*}[t]
	\centering
	\subfigure[\scriptsize P1]{
	\includegraphics[width=0.23\columnwidth]{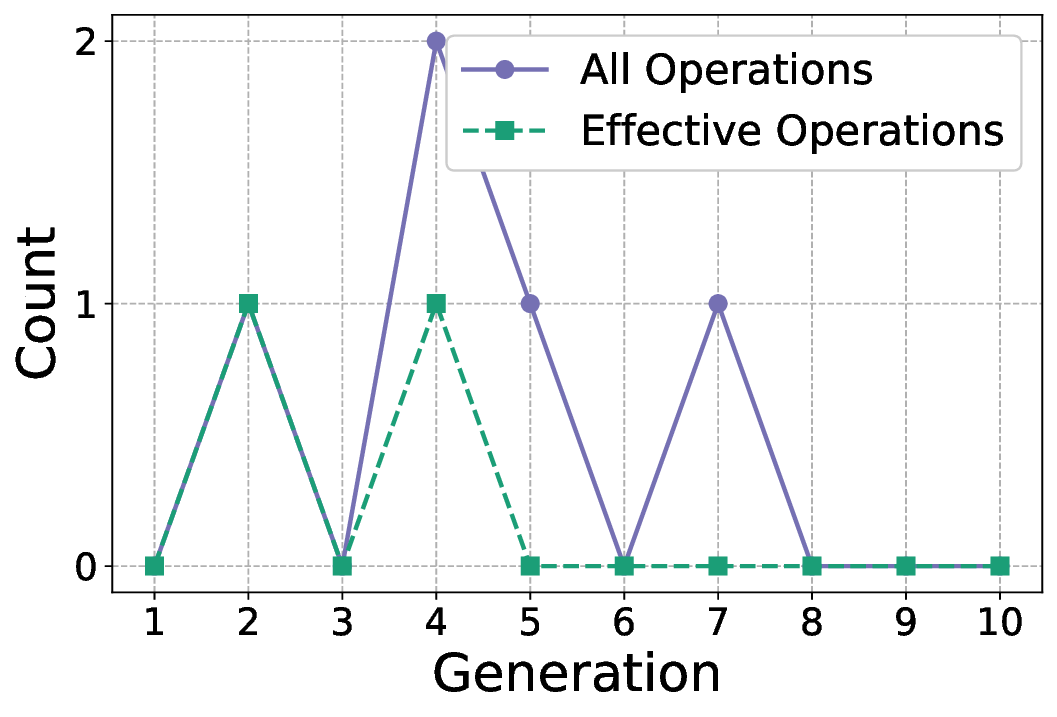}}
	\subfigure[\scriptsize P2]{
	\includegraphics[width=0.23\columnwidth]{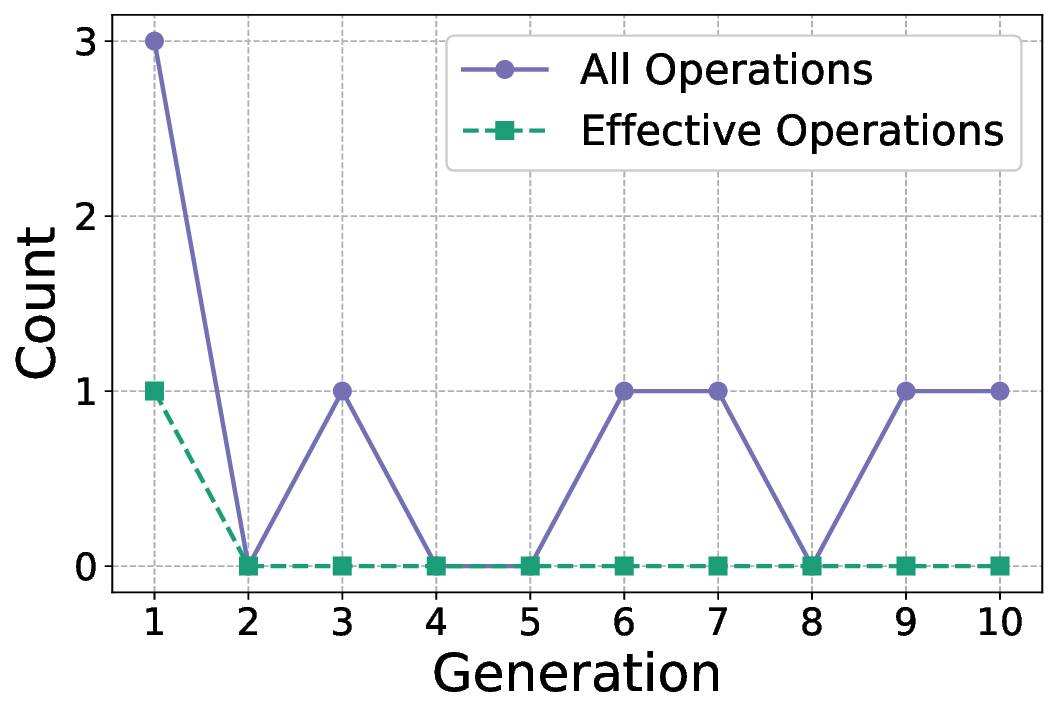}}
	\subfigure[\scriptsize P5]{
	\includegraphics[width=0.23\columnwidth]{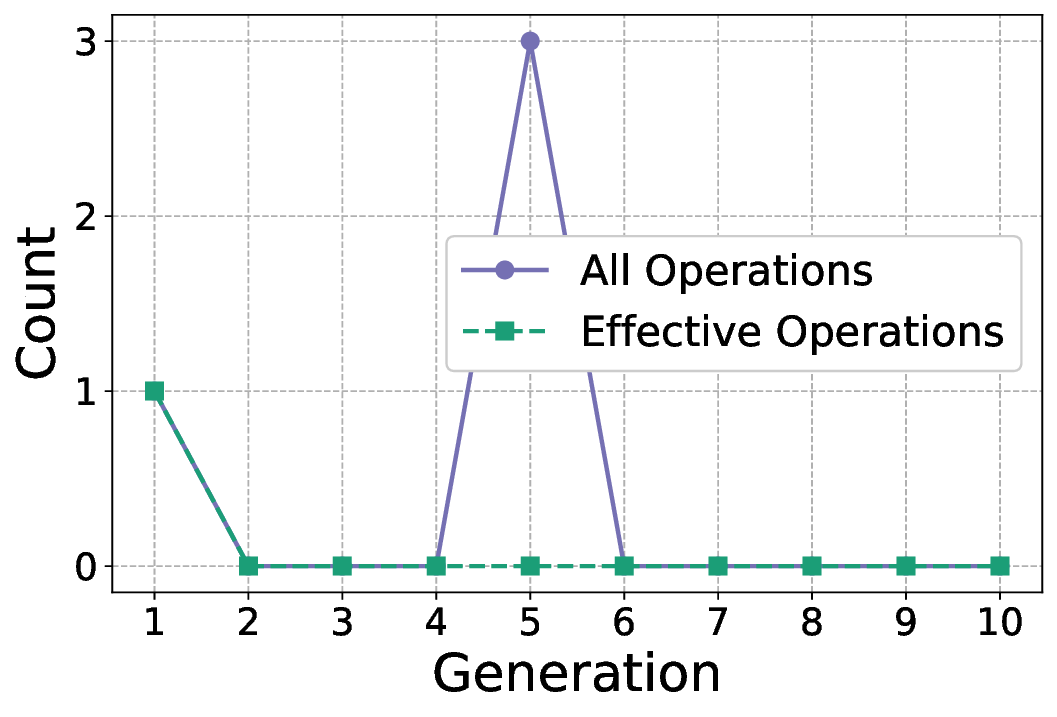}}
	\subfigure[\scriptsize P8]{
	\includegraphics[width=0.23\columnwidth]{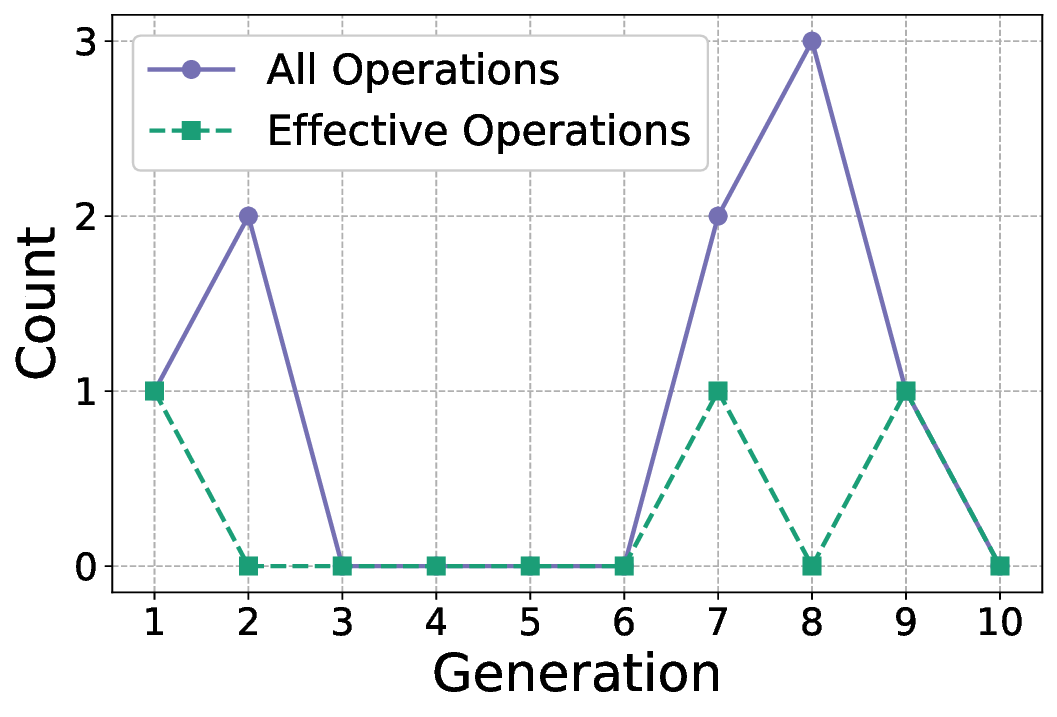}}
	\caption{Counts of \textit{improve} operations and effective \textit{improve} operations at each generation on representative instances. Y-axis: number of \textit{improve} operations and effective \textit{improve} operations; X-axis: generations.}
	\label{fig:success_rate}
\end{figure*}

The detailed behavior of the \textit{improve} operator is depicted in Fig. \ref{fig:success_rate}. In this figure, the purple solid lines denote the number of \textit{improve} operations executed in each generation, while the green dashed lines indicate the subset of operations that were effective. An effective \textit{improve} operation is defined as one that generates a KTM with higher performance than its parent. 
As can be observed, successful \textit{improve} operations predominantly occur in the early stages of the search process, where they inject useful knowledge into the population.
In P5, an effective \textit{improve} operation appears in the first generation and directly yields the best KTM of the population, as shown in Figs. \ref{fig:best_ktm}(c) and \ref{fig:success_rate}(c).
Although no effective \textit{improve} operations are observed in later generations and this best KTM is eventually replaced, the best KTMs appearing in the subsequent generations remain implicitly influenced by the earlier \textit{improve} operation.
This suggests that the knowledge introduced by the \textit{improve} operation can be propagated through other operators, thereby indirectly enhancing later KTM constructions. Note that a KTM generated by the \textit{improve} operator may not necessarily be the best at the time of its creation, yet it can still play a pivotal role in guiding the evolutionary process.
As shown in Fig. \ref{fig:success_rate}(d), \textit{improve} operations occur in generations 1 and 2, and no \textit{improve} operations are recorded in generations 3-6. During generations 1-5, the best KTMs are not linked to the \textit{improve} operation; however, the best KTM in generation 6 is produced implicitly via knowledge propagated from earlier \textit{improve} operations, as shown in Fig. \ref{fig:best_ktm}(d).
These findings highlight that although the \textit{improve} operator may not always yield the immediate best KTM, it contributes to the evolutionary dynamics by introducing self-guided knowledge that facilitates the emergence of high-performance KTMs in subsequent generations.

\subsubsection{Insights of Knowledge Acquisition}

\begin{figure*}[!b]
	\centering
	\subfigure[\scriptsize Generations 1-2 in P1]{
	\includegraphics[width=0.32\columnwidth]{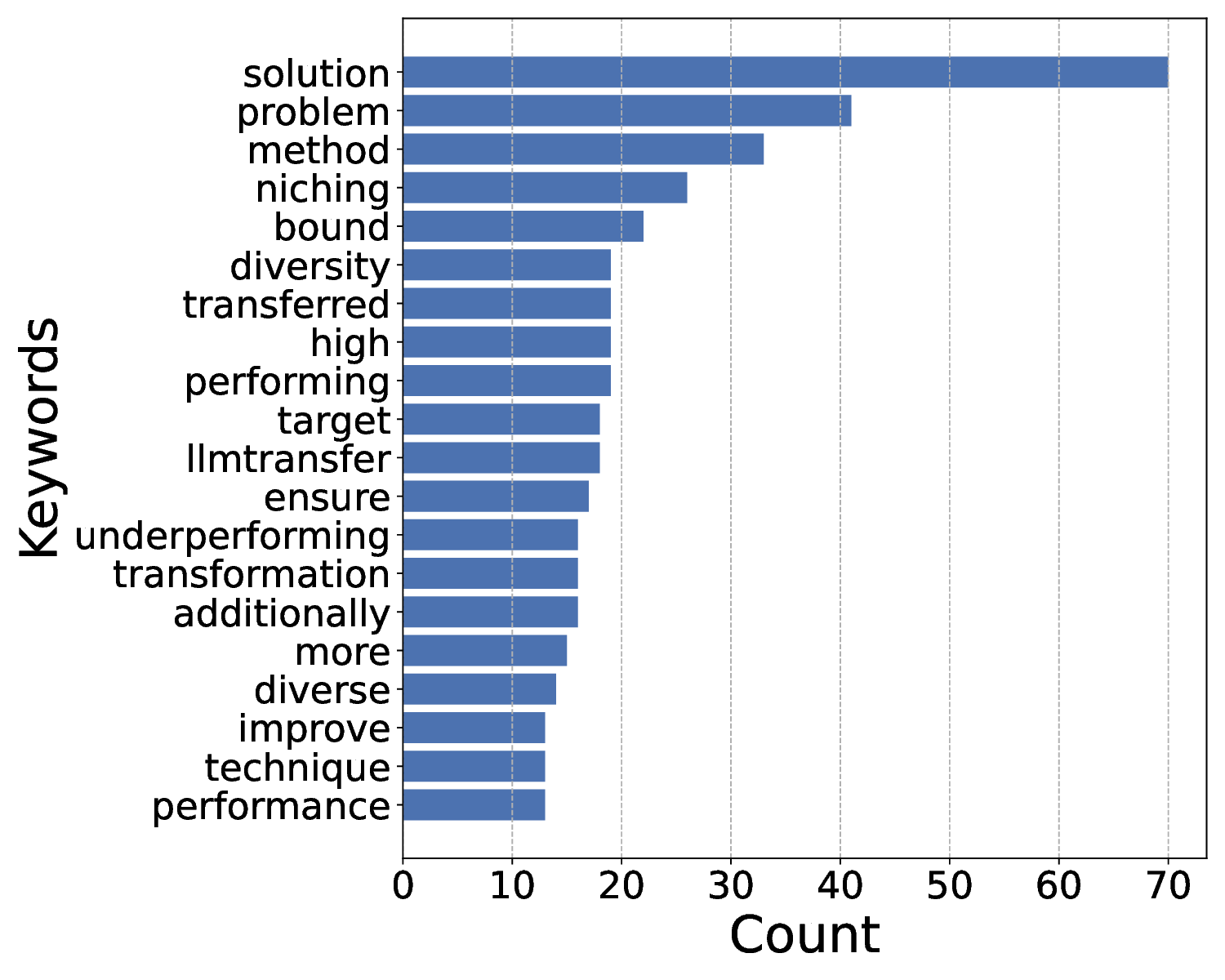}}
	\subfigure[\scriptsize Generations 3-5 in P1]{
	\includegraphics[width=0.32\columnwidth]{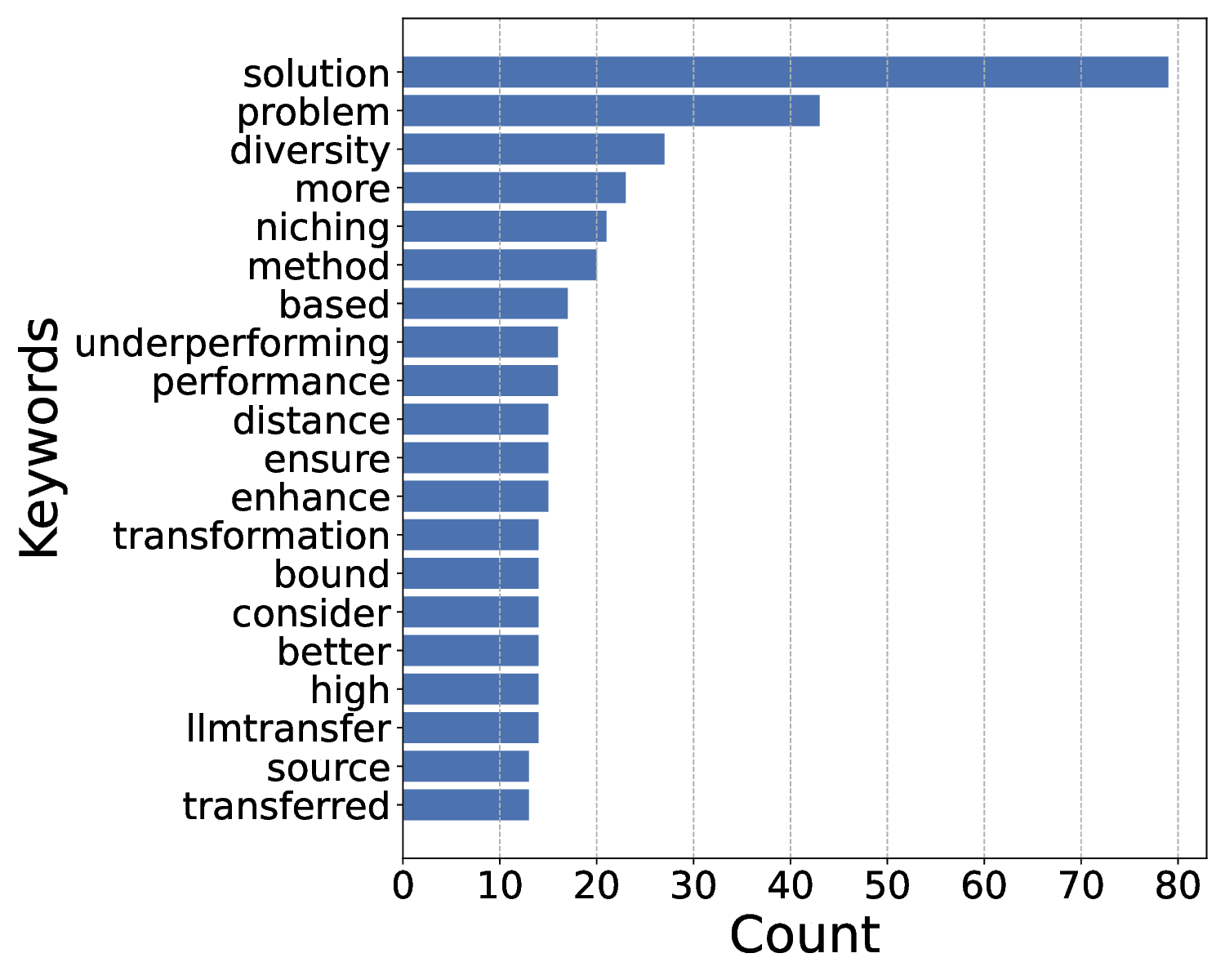}}
	\subfigure[\scriptsize Generations 6-10 in P1]{
	\includegraphics[width=0.32\columnwidth]{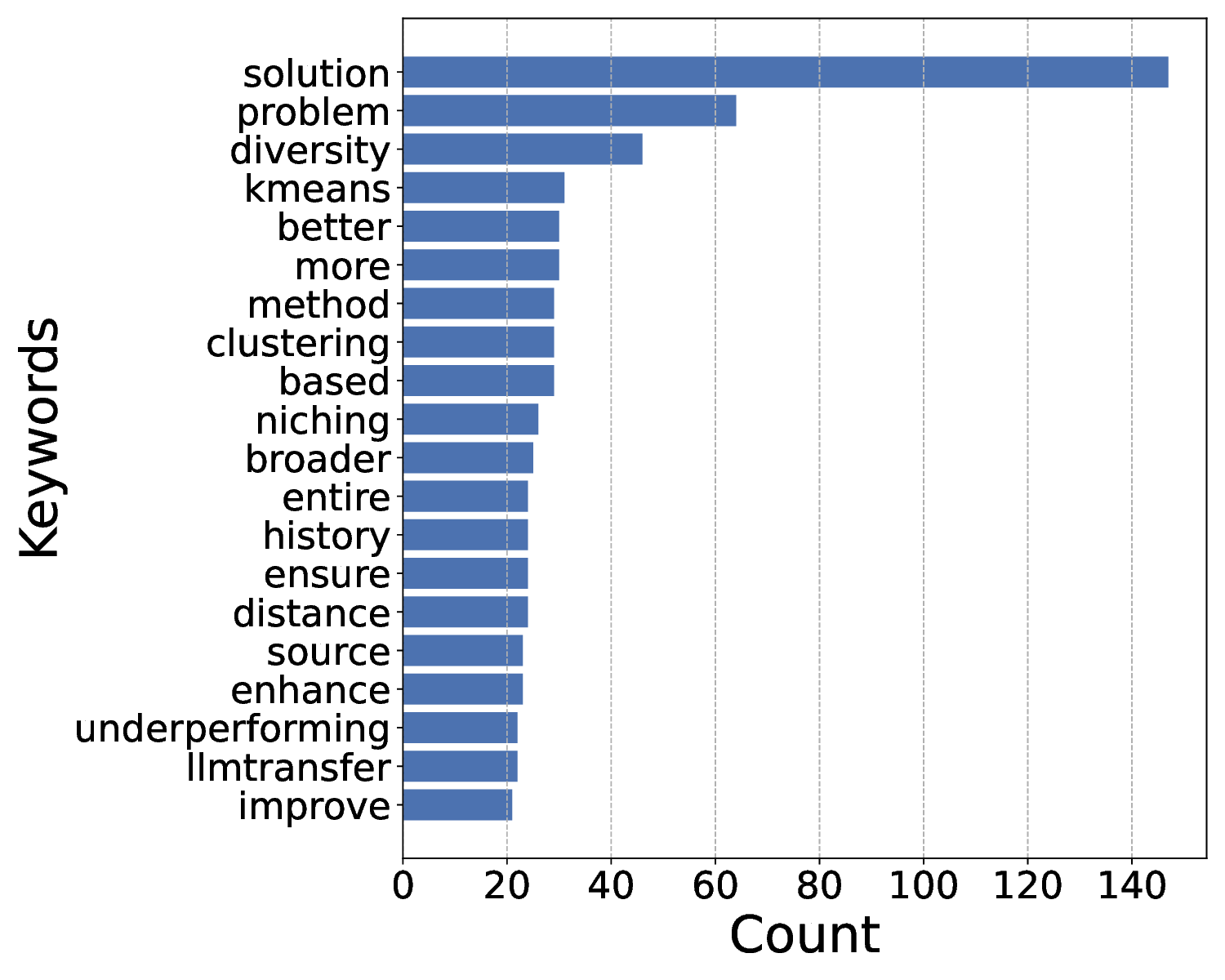}}
	\subfigure[\scriptsize Generations 1-2 in P5]{
	\includegraphics[width=0.32\columnwidth]{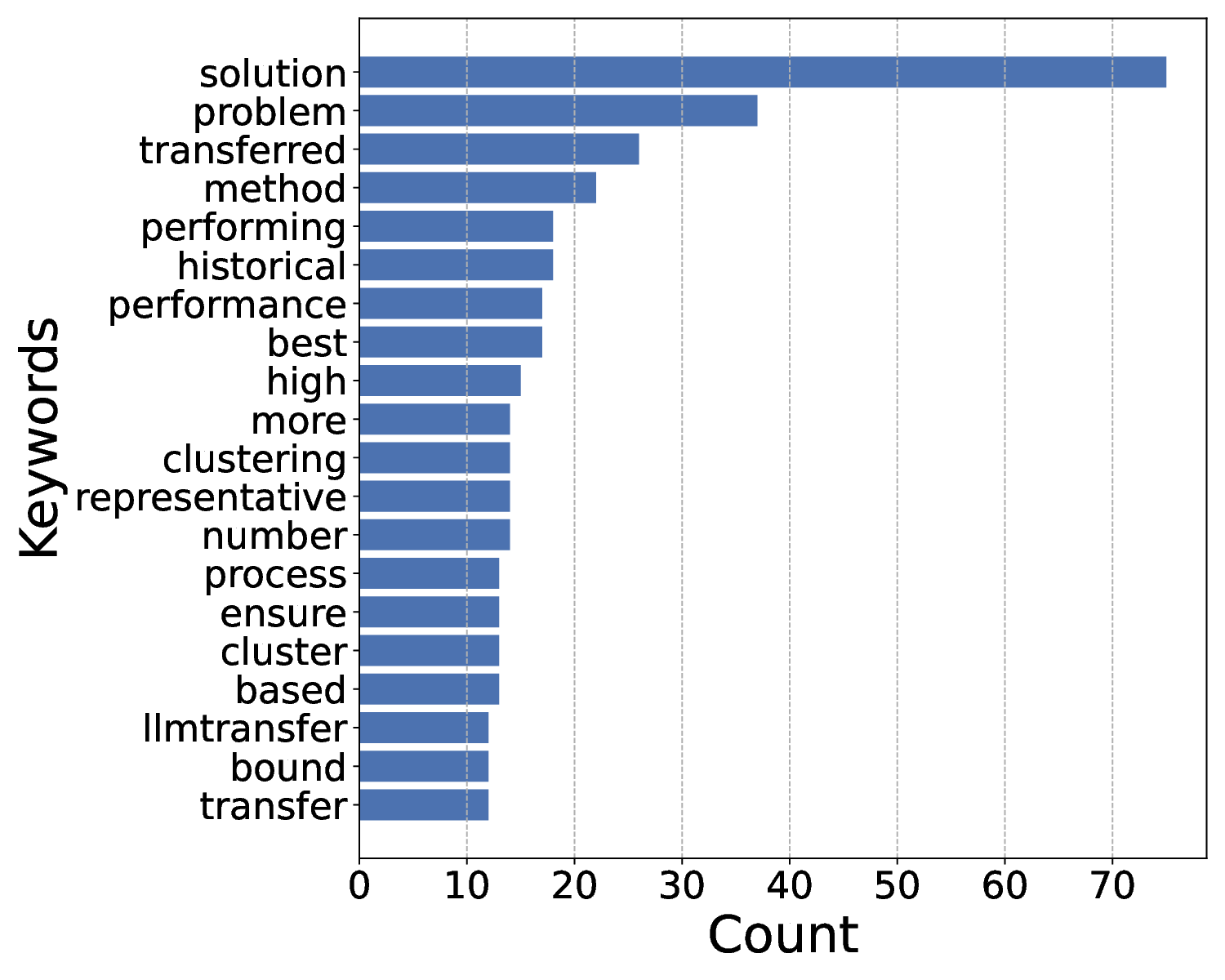}}
	\subfigure[\scriptsize Generations 3-5 in P5]{
	\includegraphics[width=0.32\columnwidth]{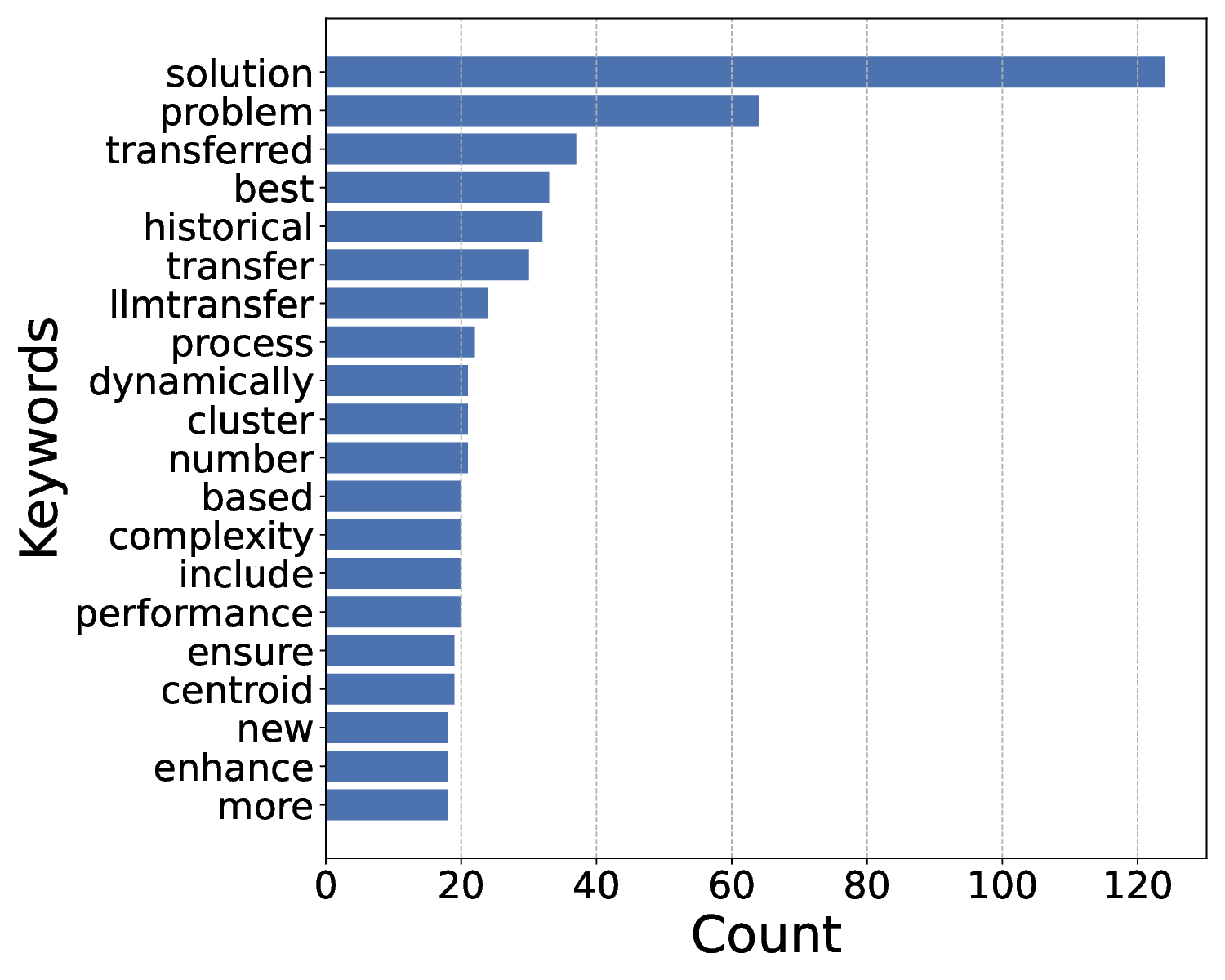}}
	\subfigure[\scriptsize Generations 6-10 in P5]{
	\includegraphics[width=0.32\columnwidth]{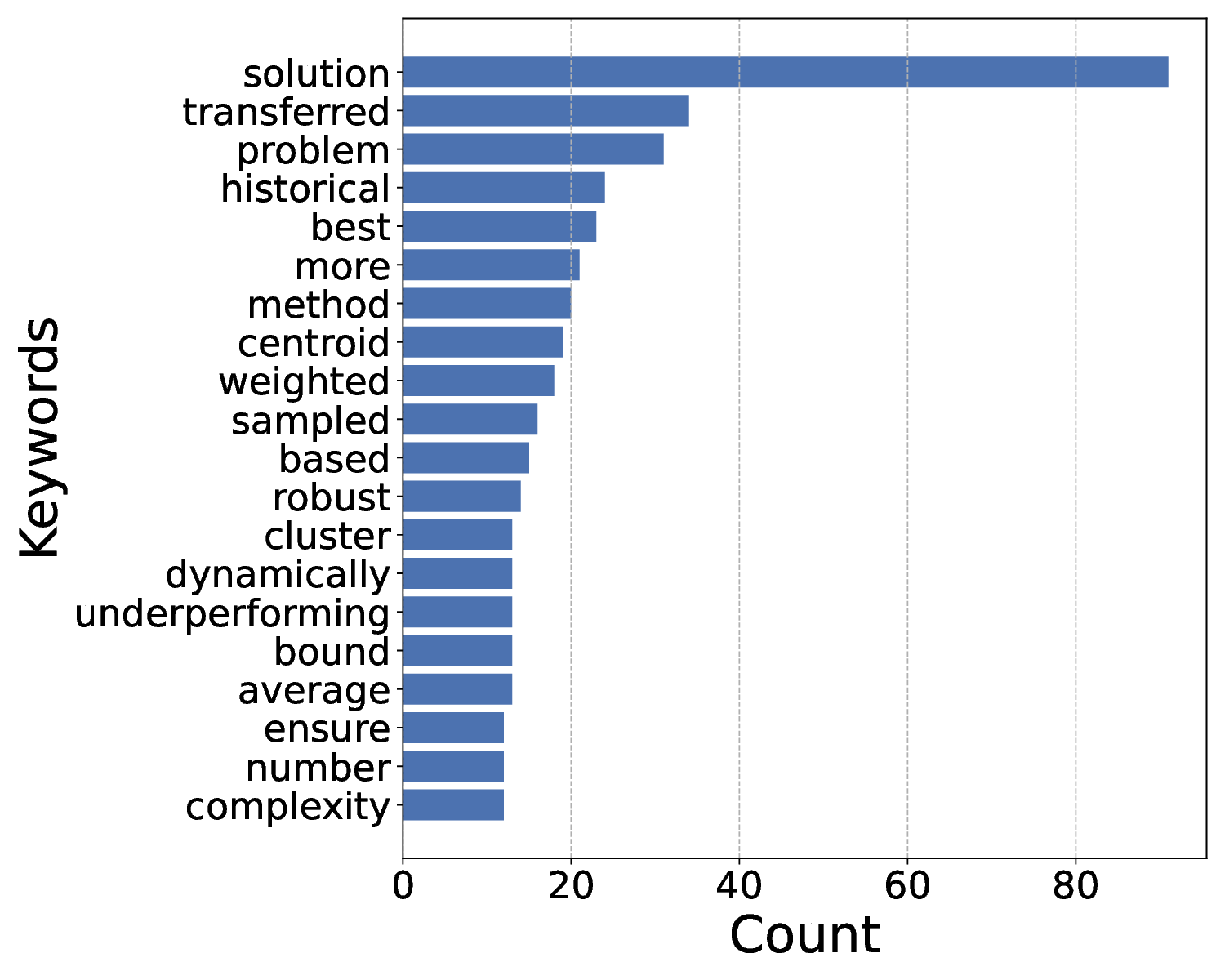}}
	\caption{The top 20 most frequent keywords in the acquired knowledge over different generation intervals on representative instances. Y-axis: keywords; X-axis: occurrence counts.}
	\label{fig:word_fre}
\end{figure*}

To investigate the evolution of the acquired knowledge throughout the search process across different scenarios, Fig. \ref{fig:word_fre} presents the most frequently occurring keywords in the acquired knowledge at different generation intervals on representative instances.
As shown in Fig. \ref{fig:word_fre}, the most recurrent keywords across all subfigures include ``solution'' and ``problem'', suggesting that these concepts constitute the core foundation of the improvement knowledge.
Moreover, the acquired knowledge shows certain differences across different EMTO scenarios, indicating that the improvement knowledge is scenario-dependent.
For example, ``niching'' remains a high-frequency keyword throughout the search process in P1, but not in P5. This suggests that the knowledge acquired in P1 places greater emphasis on niching methods \cite{li2016seeking}, whereas such an emphasis is less evident in P5.
Furthermore, keyword distributions also evolve across different stages within the same scenario.
For example, in P1, ``bound'' not only becomes less frequent as the search progresses, but also drops in its relative rank among all keywords by occurrence, as shown in Fig. \ref{fig:word_fre}(a) and (b).
Consequently, it no longer remains among the high-frequency keywords in the later stages. By contrast, ``clustering'' which is not among the prominent keywords initially, emerges as a high-frequency keyword in a later stage, as shown in Fig. \ref{fig:word_fre}(d).
This case indicates that the improvement knowledge in the early stage of P1 is mainly centered on boundary handling, while in the later stage it tends to place more emphasis on clustering-based strategies for improving the performance of KTMs.
Overall, these observations suggest that the improvement knowledge in the KTM search process is not static, but evolves with the search progress and varies across scenarios. This also underscores the utility of the knowledge acquisition mechanism for identifying and incorporating appropriate knowledge during the KTM search process under different scenarios.

\begin{figure}[htbp]
	\centering
	\includegraphics[width=0.35\columnwidth]{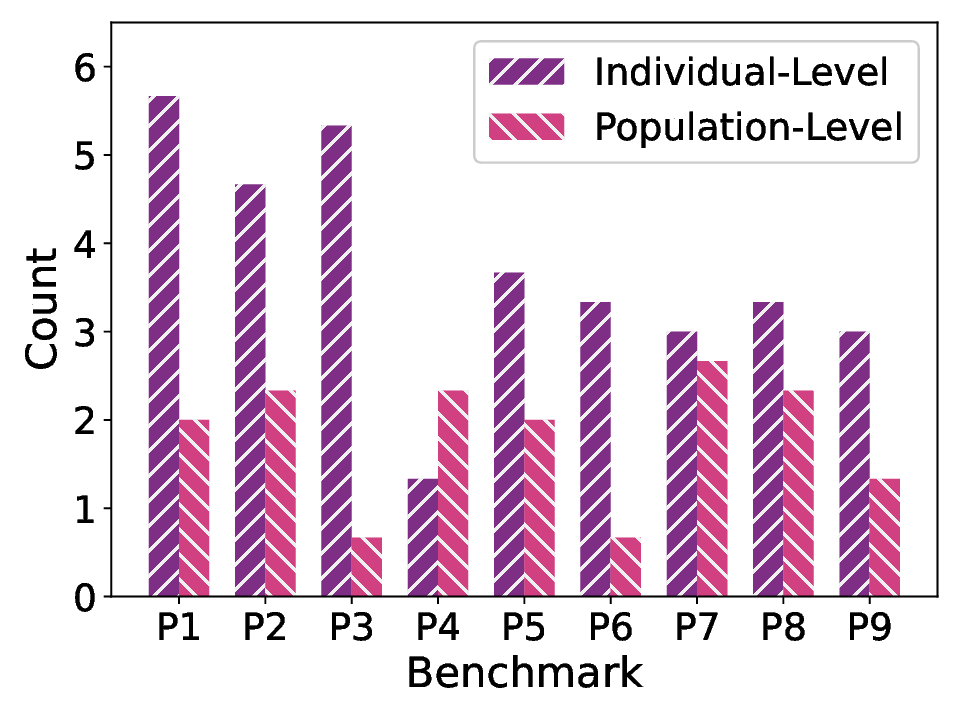}
	\caption{Mean final quantities of individual- and population-level knowledge obtained on different benchmarks over three runs. Y-axis: count of knowledge items; X-axis: benchmarks.}
	\label{fig:hint_source}
\end{figure}

To analyze the roles of individual- and population-level knowledge, Fig.~\ref{fig:hint_source} first presents the mean counts of final individual- and population-level knowledge obtained after the semantic de-duplication mechanism.
As can be observed, the final acquired knowledge contains both individual- and population-level knowledge, thereby demonstrating the validity of both acquisition mechanisms.
Moreover, the average number of acquired knowledge items across all test scenarios remains below ten, which motivates the setting of the number of most relevant knowledge entries in the \textit{improve} operator.
This setting is designed to include most acquired knowledge items, and often all of them, given that the amount of knowledge is typically limited across scenarios.
At the same time, it prevents excessive knowledge inputs in rare cases with larger knowledge sets, which could otherwise degrade the analytical effectiveness of the LLM.
Furthermore, the typically larger amount of individual-level knowledge indicates that fine-grained evolution retains a rich source of effective instance-specific information tailored to the current EMTO scenario.
Notably, the individual-level knowledge acquisition mechanism is triggered under more stringent conditions than its population-level counterpart, as it requires explicit performance improvements in parent-offspring comparisons. This further highlights the richness and practical value of individual-level knowledge.

\begin{figure}[htbp]
	\centering
	\subfigure[\scriptsize P1]{
	\includegraphics[width=0.15\columnwidth]{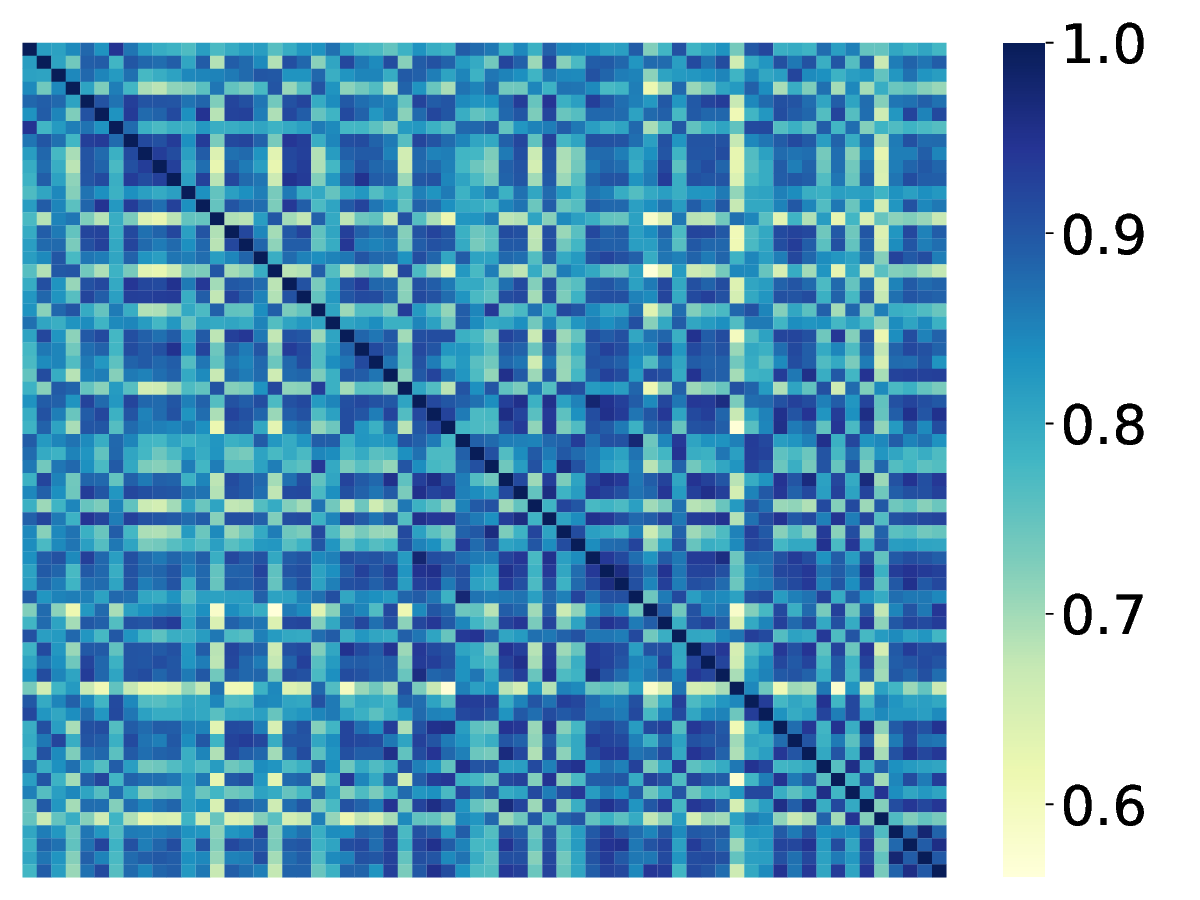}}
	\subfigure[\scriptsize P2]{
	\includegraphics[width=0.15\columnwidth]{resources/6/part/A_R1_G1.eps}}
	\subfigure[\scriptsize P5]{
	\includegraphics[width=0.15\columnwidth]{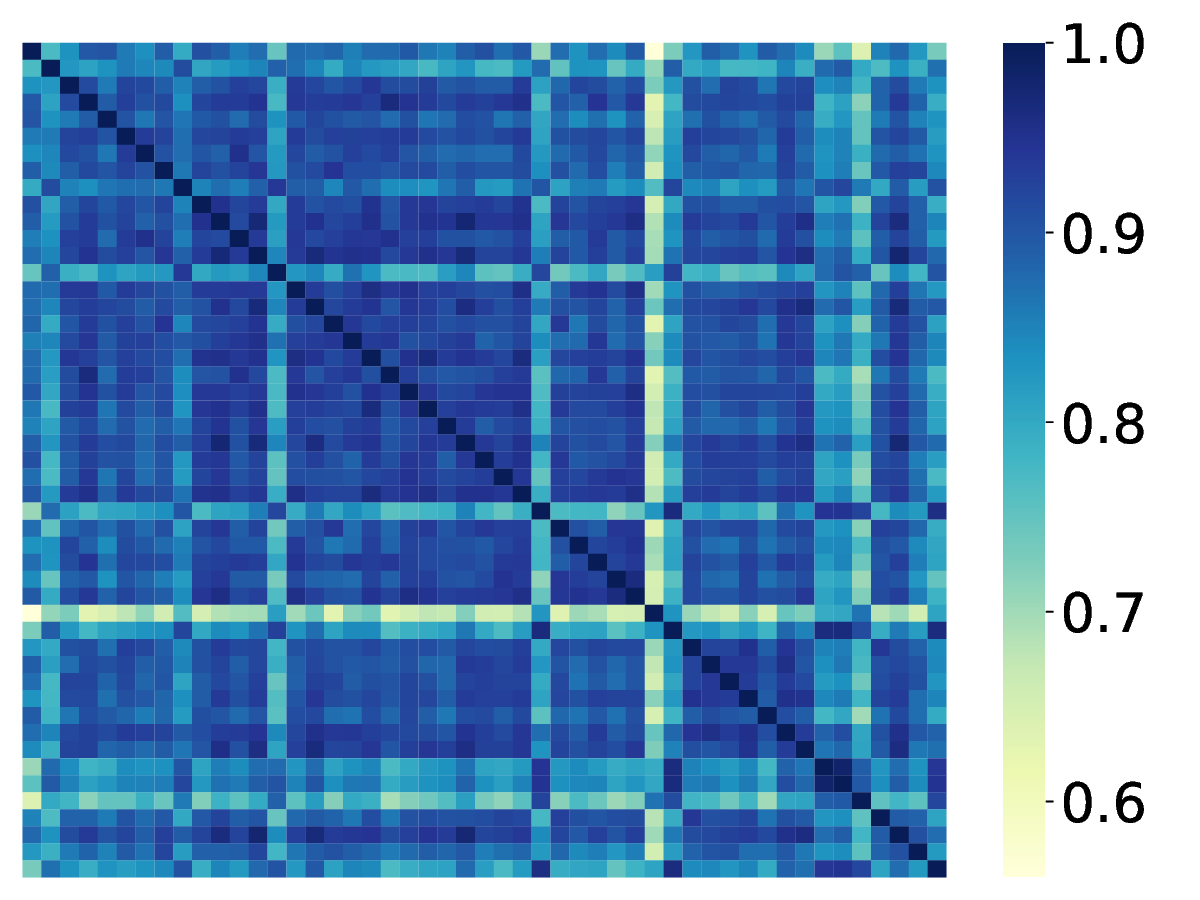}}
	\subfigure[\scriptsize P8]{
	\includegraphics[width=0.15\columnwidth]{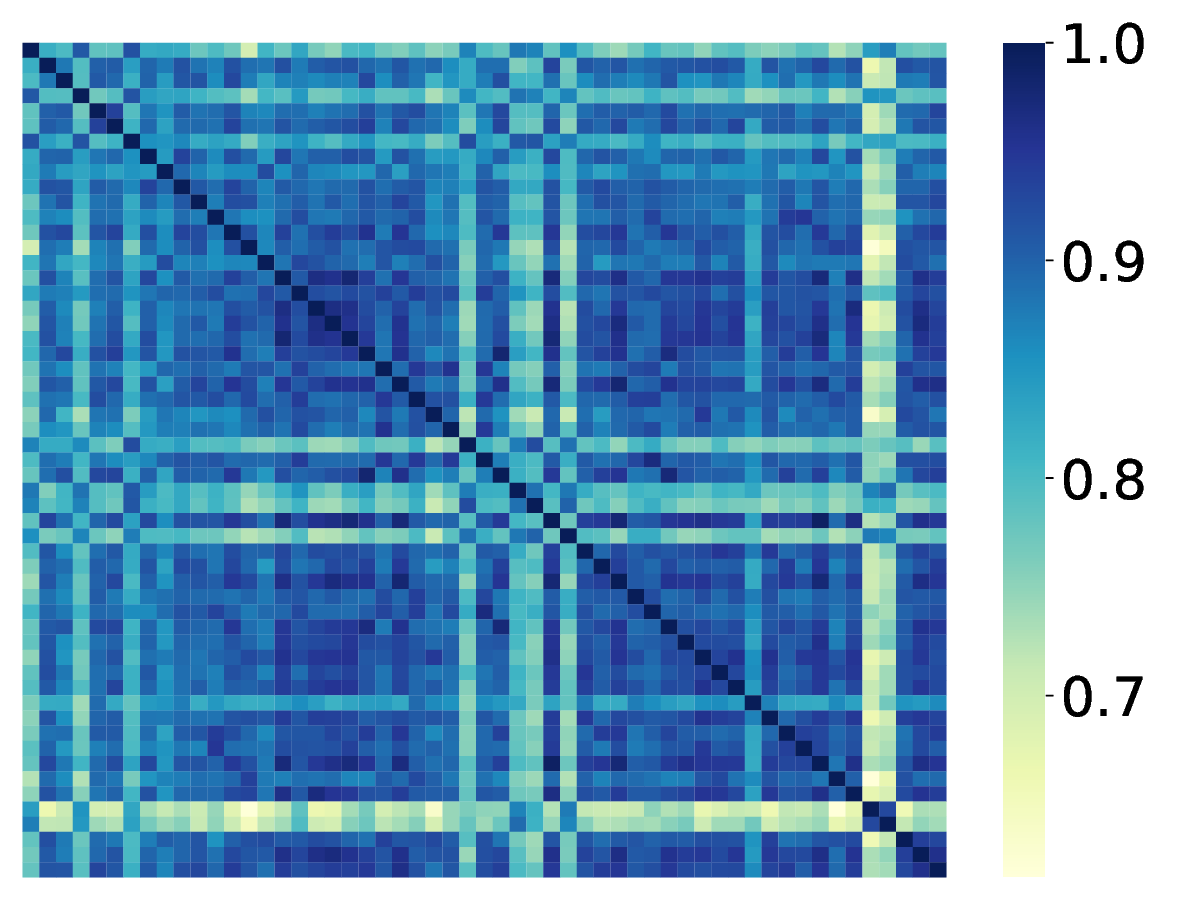}}
	\caption{Heatmap of pairwise similarities across the entire set of generated knowledge on representative instances.}
	\label{fig:sim_all}
\end{figure}

\begin{figure}[htbp]
	\centering
	\subfigure[\scriptsize P1]{
	\includegraphics[width=0.15\columnwidth]{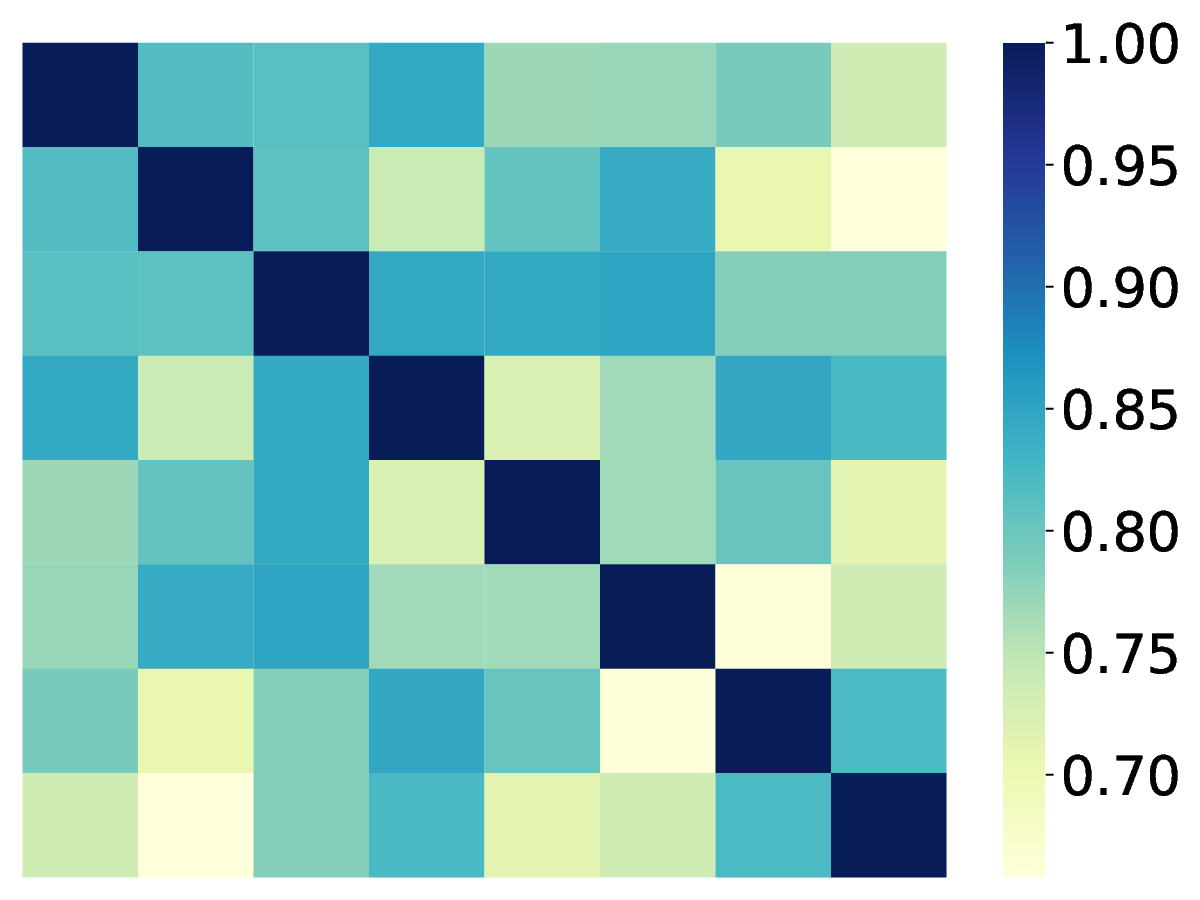}}
	\subfigure[\scriptsize P2]{
	\includegraphics[width=0.15\columnwidth]{resources/6/part/R1_G1.eps}}
	\subfigure[\scriptsize P5]{
	\includegraphics[width=0.15\columnwidth]{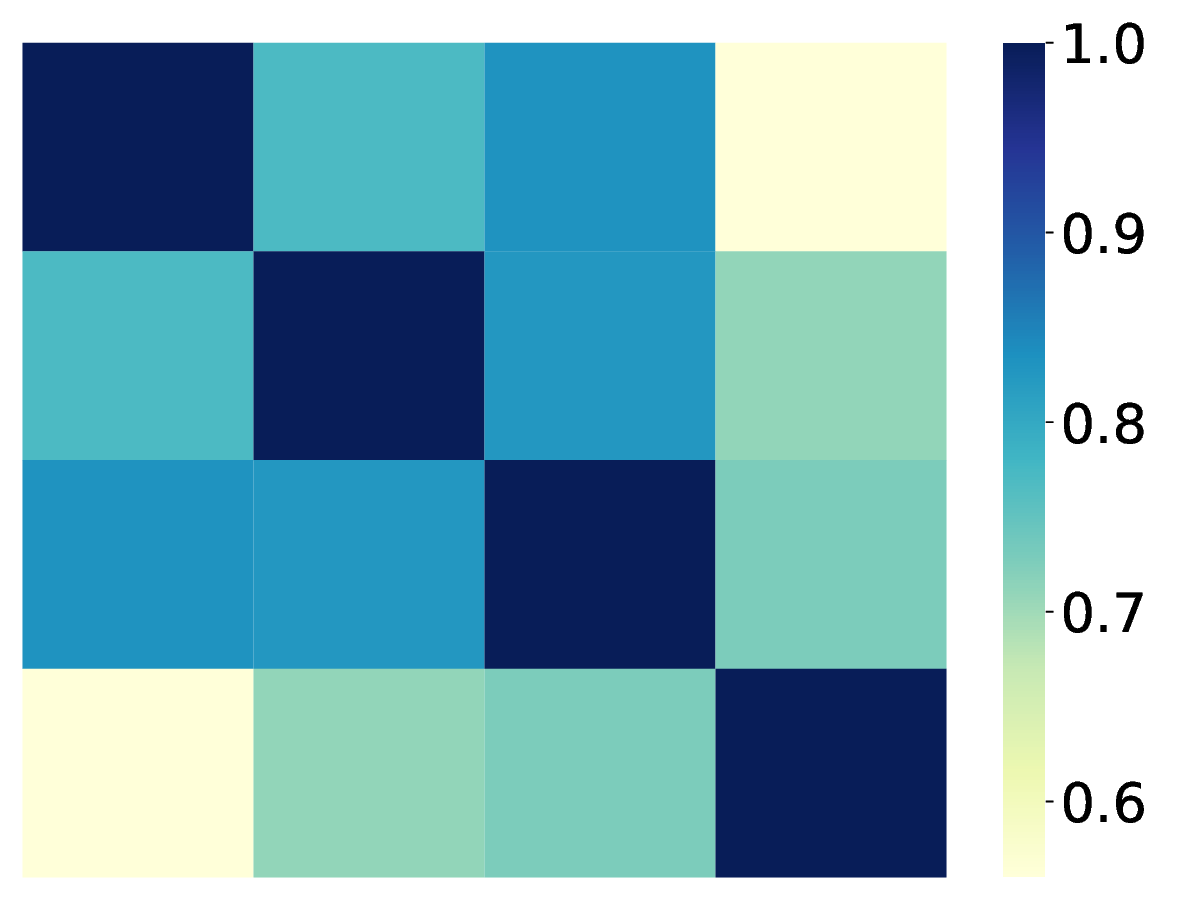}}
	\subfigure[\scriptsize P8]{
	\includegraphics[width=0.15\columnwidth]{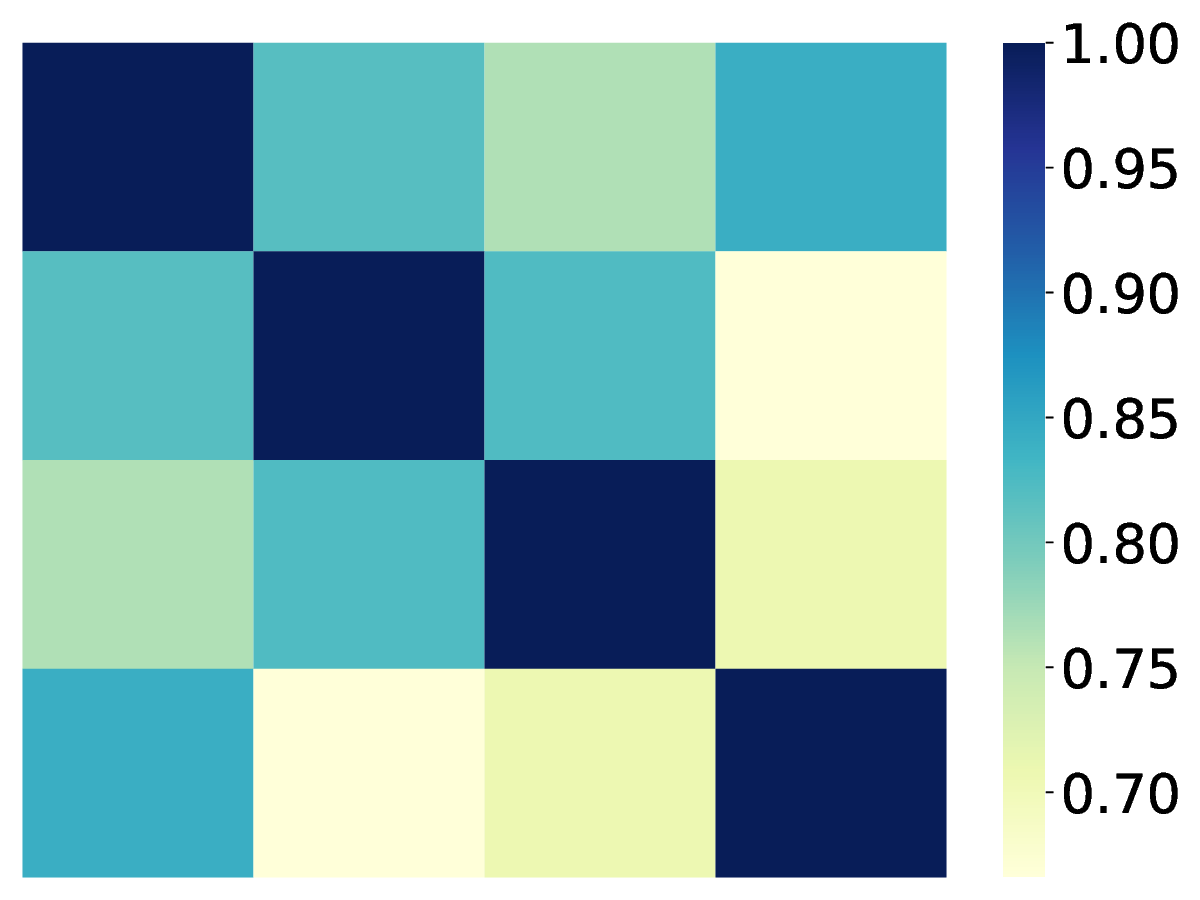}}
	\caption{Heatmap of pairwise similarities among duplicate knowledge on representative instances.}
	\label{fig:sim}
\end{figure}

To analyze the semantic de-duplication mechanism, similarity heatmaps for the full set of knowledge items and the de-duplicated set are presented in Fig. \ref{fig:sim_all} and Fig. \ref{fig:sim}, respectively.
As shown in Fig. \ref{fig:sim_all}, the LLM generates a substantial number of highly similar knowledge items.
Specifically, the total numbers of generated knowledge items in the four representative instances are 64, 59, 49, and 55, respectively.
Such duplication interferes with the effective transfer of self-guided knowledge by lengthening the input provided to the LLM and diminishing the impact of other diverse knowledge items.
After applying the semantic de-duplication mechanism, the total number of knowledge items is reduced considerably, and the remaining set exhibits more pronounced variations in similarity, as shown in Fig. \ref{fig:sim}.
This de-duplication mechanism enables the LLM to exploit self-guided knowledge more effectively during the \textit{improve} operations.

\section{Conclusion} \label{sec:conclusion}
In this study, we have proposed an LLM-assisted method, namely SKTD, to generate effective KTMs for diverse EMTO scenarios autonomously, addressing the limitations of manually designed KTMs that require substantial expertise and exhibit limited adaptability.
By utilizing LLM-based operators, SKTD is able to generate high-quality KTMs that enhance the optimization process while minimizing the need for substantial expert knowledge and human intervention.
By leveraging the design knowledge acquired during the search process, SKTD can further improve the performance of KTM generation.
Comprehensive empirical studies across EMTO scenarios with varying degrees of similarity demonstrate that the proposed SKTD algorithm outperforms both the state-of-the-art program search method and existing manually designed KTMs.

The findings of this research pave the way for autonomous knowledge transfer design in evolutionary multitasking. In the future, we would like to extend the applicability of the SKTD algorithm to a broader range of EMTO problems and investigate the integration of additional advanced techniques to further enhance the LLM's effectiveness in KTM generation. Furthermore, we plan to evaluate the potential of our algorithm in real-world EMTO applications, ensuring its versatility and effectiveness across diverse optimization scenarios.

\bibliographystyle{unsrtnat}
\bibliography{resources/LLM,resources/EMTO,resources/KTM}

\end{document}